\pdfoutput=1

\documentclass{article}
\usepackage{colm2024_conference}

\usepackage{url}
\usepackage{booktabs}
\usepackage{multirow}
\usepackage{hyperref}
\usepackage{graphicx}
\usepackage{microtype}
\usepackage{xcolor}
\usepackage{colortbl}
\usepackage{tabularx}
\usepackage{enumitem}
\usepackage{caption} 
\usepackage{float}
\usepackage{wrapfig}
\usepackage{amsmath,amsfonts,amssymb}
\usepackage{amsthm}
\usepackage{placeins}
\hypersetup{pdftitle={DMTP: A Better First Token for Block Drafting via MTP},pdfauthor={Yuhao Shen and collaborators}}

\makeatletter
\def\@fnsymbol#1{\ensuremath{\ifcase#1\or \dagger \or  \ddagger\or
   \mathsection\or  \text{*}\or \mathparagraph \or  \| \or **\or \dagger\dagger
   \or \ddagger\ddagger \else\@ctrerr\fi}}
\makeatother
\renewcommand{\thefootnote}{\fnsymbol{footnote}}

\title{DMTP: A Better First Token for Block Drafting via MTP}

\author{
\textbf{%
Yuhao Shen$^{1,2,\dagger,\ddagger}$ \quad Jianuo Huang$^{3,\dagger}$ \quad Yi Zhang$^{1}$ \quad Yang Xu$^{1}$ \quad Hao Yu$^{1}$ \\ Xin Wang$^{1}$ \quad
Yucheng Li$^{1}$ \quad Huiqiang Jiang$^{1}$  \quad Xi Lin$^{1}$ \quad Di Liu$^{1}$ \quad Jun Zhang$^{2}$ \\
Tianyu Liu$^{4}$ \quad Bo Zheng$^{1}$ \quad Dayiheng Liu$^{1}$ \quad Jianwei Zhang$^{1,*}$ \quad Cong Wang$^{2,*}$} \\
\vspace{1.5mm}
\small $^{1}$Alibaba Token Hub, Alibaba Group \quad $^{2}$ZJU \quad $^{3}$HUST \quad $^{4}$USTC\\[2mm]
\small\href{https://sylvan820.github.io/D-MTP/}{\raisebox{-0.13em}{\includegraphics[height=0.95em]{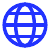}}\hspace{0.35em}Blog}\quad
\href{https://github.com/Sylvan820/sglang-dmtp}{\raisebox{-0.13em}{\includegraphics[height=0.95em]{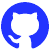}}\hspace{0.35em}GitHub}\quad
\href{https://huggingface.co/Piggy12/Qwen3.6-27B-DMTP}{\raisebox{-0.13em}{\includegraphics[height=0.95em]{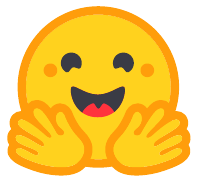}}\hspace{0.35em}Hugging Face}
}

\begin{document}
\maketitle

\footnotetext[1]{Equal contribution.}
\footnotetext[4]{Corresponding authors.}
\footnotetext[2]{Work was done during an internship at Alibaba Token Hub, Alibaba Group.}
\renewcommand{\thefootnote}{\arabic{footnote}}
\setcounter{footnote}{0}

\begin{abstract}
Speculative decoding accelerates Large Language Model (LLM) inference by verifying lightweight draft proposals in parallel. Parallel block drafters such as DFlash generate multiple candidates in a single pass, yet strict left-to-right verification discards the entire block whenever the first candidate is rejected. Conversely, native Multi-Token Prediction (MTP) modules trained alongside the target model offer strong first-token accuracy and out-of-distribution generalization, but incur serial latency when extended across multiple steps.  To address these limitations, we introduce \textbf{DMTP}, a cooperative speculative decoding framework where native MTP initiates each block and DFlash generates the remaining continuation in parallel. Specifically, MTP proposes the first token and injects its hidden state into DFlash via a lightweight residual on that token's embedding to leverage extra predictive information from MTP. Conditioning parallel drafting on a concrete first token both inherits MTP's first-token accuracy and shifts the remaining block one step forward, making subsequent tokens easier to predict. Across short- and long-context benchmarks on Qwen3.6-27B, Qwen3.6-35B-A3B, and Qwen3.8-Next-120B, DMTP consistently outperforms DFlash and DFlash2 while surpassing Native MTP on nearly all workloads. It improves accepted length by up to 35.64\% over DFlash2 on long-context workloads and boosts workload-serving throughput over DFlash by 17.10\%--23.75\%. Moreover, DMTP is orthogonal to existing MTP and DFlash optimizations, with further gains from checkpoint adaptation and joint fine-tuning.
\end{abstract}

\section{Introduction}
\label{sec:dmtp-intro}

Large Language Models (LLMs) generate text autoregressively, requiring a full forward pass for each generated token. This sequential bottleneck makes decoding latency scale with output length and leaves GPU compute underutilized. Speculative decoding~\citep{leviathan2023fast,chen2023accelerating} alleviates this limitation by using a lightweight draft model to propose $K$ candidate tokens and verifying them in a single parallel pass of the target model. Because verification accepts tokens strictly from left to right and stops at the first rejection, practical speedup depends on both low drafting overhead and long accepted prefixes. In particular, the first draft token is critical: once it is rejected, the entire remaining block is discarded no matter how accurate the following tokens are.

Parallel block drafters such as DFlash~\citep{chen2026dflash} generate all $K$ candidates in a single forward pass, avoiding sequential drafting overhead. Recent follow-ups such as Domino~\citep{domino}, DSpark~\citep{cheng2026dspark}, and DFlash2~\citep{dflash2} further introduce intra-block autoregressive refinements to alleviate accuracy decay at later positions. However, this line of work largely overlooks first-token accuracy, and because parallel drafters are trained via supervised fine-tuning (SFT) on limited data, they still suffer from weak first-token accuracy and poor out-of-distribution (OOD) generalization. To bridge this gap, we turn to native Multi-Token Prediction (MTP)~\citep{gloeckle2024bmtp,deepseekv3,qwen35}: trained jointly with the target model on pretraining-scale corpora, it offers strong first-token accuracy and broad domain coverage. On Qwen3.6-35B-A3B Agent workloads, native MTP reaches 90.17\% first-token acceptance versus 79.53\% for DFlash. Yet generating an entire block with native MTP requires sequential steps whose latency grows with block length. This contrast raises a natural question: can we reuse native MTP to produce a stronger first candidate, and let a parallel drafter complete the rest of the block conditioned on it?

\begin{figure}[!t]
\centering
\includegraphics[width=\linewidth]{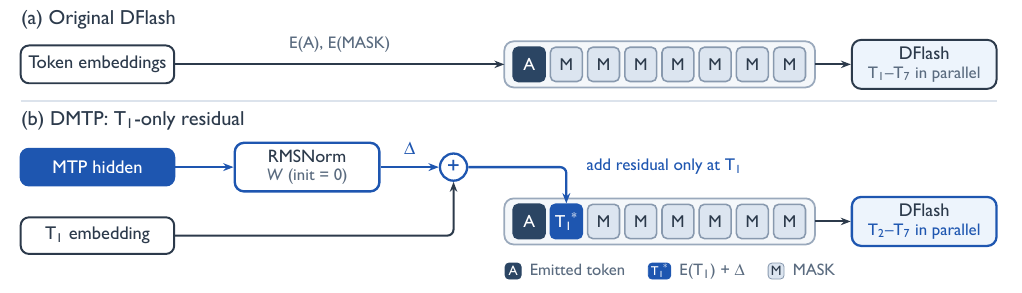}
\caption{\textbf{Architectural comparison between DFlash and DMTP.} Original DFlash predicts all seven candidates in parallel from the committed anchor token $A$. DMTP delegates the first candidate $T_1$ to native MTP and predicts $T_2$--$T_7$ in parallel. Both $A$ and $T_1$ serve as dual anchors, while MTP's normalized hidden state enters only through a zero-initialized residual on the $T_1$ embedding, leaving DFlash's target-context KV interface unchanged.}
\label{fig:dmtp-intro-interfaces}
\end{figure}

We introduce \textbf{DMTP}, which lets DFlash reuse native MTP's pretrained capabilities while retaining parallel continuation (Figure~\ref{fig:dmtp-intro-interfaces}). In each round, MTP takes the last committed token $A$ and target features to propose the first candidate $T_1$ and its hidden state. DFlash then takes the dual-anchor prefix $[A, T_1]$ followed by $K-1$ mask tokens ($K=7$ in our main setup) to predict $T_2,\ldots,T_K$ in one parallel pass. Orthogonal to existing MTP and parallel-drafting optimizations (such as DFlash and DFlash2), DMTP offers three key advantages:

\paragraph{1) Higher first-token accuracy and OOD generalization.}
Reusing native MTP to propose $T_1$ brings its pretraining-scale first-token accuracy into parallel drafting without incurring multi-step sequential latency. On Qwen3.6-35B-A3B, DMTP raises first-token acceptance over DFlash from 71.00\% to 81.81\% on Code and from 59.33\% to 71.79\% on SWE (+10.81 and +12.46 percentage points). Because MTP is pretrained alongside the target model on broad corpora, DMTP generalizes reliably to OOD workloads beyond DFlash's SFT data while still supporting joint fine-tuning on specialized domains.

\paragraph{2) Prefix-shifted parallel continuation.}
Providing $T_1$ upfront shifts DFlash's parallel window one step forward so that $T_2$ occupies what was originally DFlash's first predicted slot. As a result, $T_2,\ldots,T_K$ are learned more thoroughly during training and drafted one step closer to a concrete prefix at inference, lifting acceptance across all subsequent positions. Even under matched first-token accuracy, this shift alone increases accepted length by 5.97\%--9.90\%.

\paragraph{3) Lightweight MTP hidden-state injection.}
Normalizing and adding MTP's hidden state to the $T_1$ embedding via a lightweight residual lets DFlash leverage extra predictive cues beyond a discrete token ID, without adding sequence tokens or changing its KV interface.

We evaluate DMTP across short- and long-context benchmarks as well as multi-GPU serving on Qwen3.6-27B, Qwen3.6-35B-A3B, and Qwen3.8-Next-120B. On Qwen3.6-35B-A3B, DMTP increases accepted length by 10.68\%--28.25\% over DFlash and boosts four-GPU serving throughput in SGLang~\citep{sglang} by 17.10\%--23.75\%. It similarly outperforms DFlash, DFlash2 (by up to 35.64\% on held-out long-context suites), and four-step Native MTP across every benchmark on Qwen3.6-27B. Scaling to Qwen3.8-Next-120B, lightweight adaptation of an existing DFlash2 checkpoint improves accepted length by 5.28\%--16.76\% over DFlash2 across all nine workloads while surpassing Native MTP on seven of them. In summary, our main contributions are threefold:
\begin{itemize}[leftmargin=5mm,itemsep=2pt,topsep=2pt]
    \item \textbf{First-token limitations and prefix-shift gains.} We identify DFlash's first-token accuracy and OOD generalization as bottlenecks and show that reusing pretrained MTP can alleviate them. Where MTP has a clear first-token advantage, DMTP improves accepted length over DFlash by 10.68\%--28.25\% on Qwen3.6-35B-A3B. We further find that prefix shifting improves accepted length by 5.97\%--9.90\% over hidden-only conditioning, even when MTP and DFlash have similar first-token accuracy.
    \item \textbf{Combining MTP with parallel drafting.} We propose DMTP, which pairs native MTP first-token initiation with parallel DFlash continuation via a zero-initialized $T_1$ residual. Leaving both backbones and the target KV interface intact, DMTP is orthogonal to existing MTP and DFlash optimizations and supports scratch training, checkpoint adaptation, and joint fine-tuning.
    \item \textbf{Evaluation across models and workloads.} We evaluate DMTP on short- and long-context benchmarks across models ranging from 27B to 120B parameters. DMTP improves accepted length over DFlash2 by up to 35.64\% and surpasses four-step native MTP on nearly all workloads, while increasing workload-serving throughput in SGLang by 17.10\%--23.75\%.
\end{itemize}

\section{Background}
\label{sec:dmtp-background}

\begin{figure}[!t]
\centering
\includegraphics[width=0.92\linewidth]{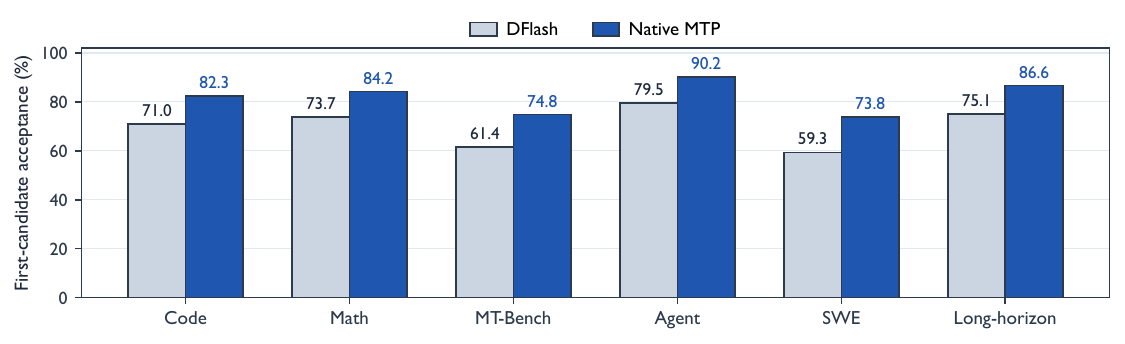}
\caption{\textbf{First-token acceptance rate of native MTP versus DFlash.} Evaluated on identical Qwen3.6-35B-A3B prompts across six workloads at temperature $1.0$, native MTP consistently achieves higher first-token acceptance than DFlash. Qwen3.6-35B DFlash is trained via SFT on a 32K-context task mixture.}
\label{fig:dmtp-first-token}
\end{figure}

\subsection{Speculative Decoding}
Speculative decoding~\citep{leviathan2023fast,chen2023accelerating} accelerates LLM inference with a lightweight draft model. In each decoding round, the drafter proposes $K$ candidate tokens $T_1,\ldots,T_K$, and the target model verifies all $K$ tokens in a single parallel forward pass, accepting the longest prefix that matches its distribution.

Under exact rejection sampling, let $p_i$ and $q_i$ denote the target and draft distributions at step $i$ given the preceding tokens. Each candidate $T_i$ is accepted from left to right with probability $\min(1,\,p_i(T_i)/q_i(T_i))$. At the first rejection at step $i$, all remaining candidates $T_{i+1},\ldots,T_K$ are discarded and a replacement token is resampled from the residual distribution proportional to $[p_i-q_i]_+$. If all $K$ candidates are accepted, the target model samples an extra bonus token from $p_{K+1}$, preserving the exact target distribution. In addition to exact rejection sampling, production serving engines and public benchmarks widely use \emph{target-only sampling}, which compares draft tokens directly against tokens sampled from the target model (reducing to exact matching under greedy decoding) without storing full draft distributions $q_i$. Throughout our experiments, we group comparisons by verification protocol so that all methods within each block share the exact same verification rule.

\subsection{Drafter Architectures: Parallel Block Drafting and Native MTP}
\paragraph{Parallel block drafting.}
Parallel drafters such as DFlash predict all $K$ candidate tokens in a single forward pass~\citep{chen2026dflash}. Let $A$ denote the anchor token, i.e., the last token committed in the previous verification round. Given $[A,\mathrm{MASK},\ldots,\mathrm{MASK}]$ with $K$ mask tokens, DFlash uses multi-layer target KV features and bidirectional attention to predict the block. All positions are predicted before $T_1$ is sampled, so later tokens cannot condition on a concrete first candidate. Domino, DSpark, and DFlash2 improve intra-block dependencies and reduce later-position accuracy decay~\citep{domino,cheng2026dspark,dflash2}. These improvements complement the question of first-token accuracy and generalization studied here.

\paragraph{Native MTP.}
Multi-Token Prediction (MTP) trains auxiliary heads or lightweight transformer layers alongside the target model during pretraining to predict future tokens from target hidden states~\citep{gloeckle2024bmtp,deepseekv3,bebop}. Native MTP uses aligned target hidden states and the previous token to predict the next token, benefiting from broad pretraining coverage. However, generating a multi-token block requires sequential steps, each depending on the preceding sampled token, so drafting latency grows with block length. DMTP uses just one MTP step to supply the first candidate, then relies on DFlash for low-cost parallel continuation.

\subsection{Prefix Survival and Throughput Trade-offs}
Let $N\in\{0,\ldots,K\}$ be the number of draft tokens accepted in a verification round. By the tail-sum identity, the expected accepted draft length equals the sum of cumulative acceptance probabilities:
\begin{equation}
\mathbb E[N]=\sum_{i=1}^{K}\Pr(N\geq i).
\label{eq:dmtp-survival}
\end{equation}
Because every term $\Pr(N\geq i)$ requires $T_1$ to be accepted first, first-token accuracy gates the entire block: improving $T_1$ lifts cumulative acceptance across all downstream positions, whereas gains at later positions take effect only if every earlier token passes verification.

Unless generation ends at EOS or reaches the output limit, appending the target model's correction or bonus token advances the sequence by $L=N+1$ tokens per round. Let $T_{\mathrm{draft}}$ and $T_{\mathrm{verify}}$ denote the wall-clock latencies of drafting and target verification. The amortized latency per generated token is then $(T_{\mathrm{draft}} + T_{\mathrm{verify}})/L$. Increasing $L$ therefore yields real serving speedups only when the extra drafting cost stays small relative to the saved target passes, motivating us to evaluate both acceptance length and end-to-end serving throughput.

\section{Motivating Observations}
\label{sec:dmtp-motivation}

\begin{figure}[!t]
\centering
\includegraphics[width=\linewidth]{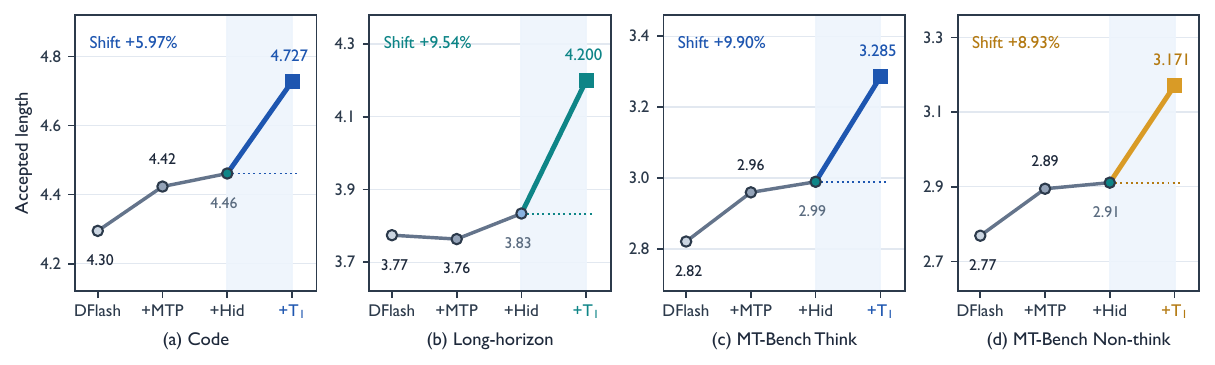}
\caption{\textbf{How the first token, hidden state, and prefix shift improve drafting on Qwen3.7-Turbo.} MTP's first-token acceptance is only 1--2 percentage points higher than DFlash's. Even with this small gap, giving DFlash the first token $T_1$ as input (\emph{Hidden + $T_1$}) increases per-verify accepted length by 5.97\%--9.90\% compared with using MTP's hidden state alone (\emph{Hidden-only}). Both trained variants use the same number of training updates.}
\label{fig:dmtp-shift-motivation}
\end{figure}

Beyond reducing intra-block accuracy decay, we investigate two complementary questions: (1) can DFlash reuse native MTP's first-token accuracy and broad domain coverage, and (2) does conditioning parallel drafting on a concrete first token still improve block acceptance when their first-token accuracies are comparable? Both ask how to improve parallel drafting without a full sequential MTP rollout.

\subsection{Pretrained First-Token Capacity}
\label{sec:dmtp-first-motivation}
Figure~\ref{fig:dmtp-first-token} compares the standalone first-token acceptance rates of native MTP and DFlash on Qwen3.6-35B-A3B across six workloads (Code, Math, MT-Bench, Agent, SWE, and Long-horizon) on identical prompts at temperature $1.0$. Here, DFlash is trained via SFT on a 32K-context task mixture, whereas native MTP uses the pretrained module shipped directly with the target checkpoint.

Across all six workloads, native MTP consistently achieves higher first-token acceptance than DFlash (for example, 82.32\% versus 71.00\% on Code). Because native MTP is trained jointly with the target model on pretraining-scale corpora, it aligns more closely with the target distribution and generalizes substantially better to out-of-distribution tasks than a lightweight drafter trained via SFT on limited data (further evaluated on held-out domains in Section~\ref{sec:dmtp-scratch}). Reusing native MTP for the first token spares the parallel drafter from relearning first-token prediction under a tight data and compute budget.

\subsection{Prefix Shifting Under Matched First-Token Quality}
\label{sec:dmtp-ablation}
While a more accurate first token naturally improves block acceptance, the first-token gap between native MTP and DFlash varies across domains and training scales. To test whether cooperative drafting remains effective even when first-token accuracy is nearly tied, we run a controlled progressive ablation on Qwen3.7-Turbo, where native MTP leads DFlash at $T_1$ by only 1--2 percentage points. Starting from standalone \emph{DFlash}, we evaluate three cumulative variants:
\begin{itemize}[leftmargin=5mm,itemsep=2pt,topsep=3pt]
    \item \emph{+ MTP first}: replaces DFlash's first token with MTP's prediction at inference time, while keeping DFlash's unshifted parallel predictions for $T_2,\ldots,T_K$.
    \item \emph{Hidden-only}: adds MTP's hidden state to DFlash during training and uses MTP's proposal as the first output token at inference. DFlash itself does not receive the $T_1$ token embedding and continues to predict the unshifted block.
    \item \emph{Hidden + $T_1$}: feeds the sampled token $T_1$ directly into DFlash's input and shifts parallel prediction one step forward to start at $T_2$ under the same training budget, isolating the effect of prefix shifting with the same first-token predictor and MTP hidden-state information.
\end{itemize}

As shown in Figure~\ref{fig:dmtp-shift-motivation}, simply plugging in MTP's first token without adapting DFlash brings only modest gains on three tasks and a slight drop on Long-horizon, since DFlash's unconditioned predictions for $T_2,\ldots,T_K$ can conflict with the externally sampled $T_1$. Injecting MTP hidden features alone adds a small further gain of 0.02--0.07 tokens per round. By contrast, feeding $T_1$ into DFlash and shifting parallel prediction to start at $T_2$ adds 0.26--0.37 accepted tokens, a 5.97\%--9.90\% relative gain over \emph{Hidden-only}. The intuition is straightforward: under this shift, DMTP's second position $T_2$ occupies what was originally DFlash's first predicted slot, and every subsequent position $T_i$ ($i\geq 2$) moves one slot closer to the known prefix. During training, this frees DFlash's highest-weight early slots from predicting $T_1$ so that $T_2,\ldots,T_K$ are learned more thoroughly. During inference, conditioning directly on the concrete $T_1$ turns each $T_i$ into an easier $(i-1)$-step prediction, lifting acceptance rates across all subsequent positions even when first-token accuracy is matched.

\paragraph{Architectural takeaway.}
Together, these findings suggest using MTP only for $T_1$ and its hidden state, then shifting DFlash to generate $T_2,\ldots,T_K$ in parallel. This reuses MTP's pretrained capabilities without multi-step sequential latency and complements improvements to DFlash's block-internal architecture and training.

\section{Architecture}
\label{sec:dmtp-method}

\begin{figure}[!t]
\centering
\includegraphics[width=\linewidth]{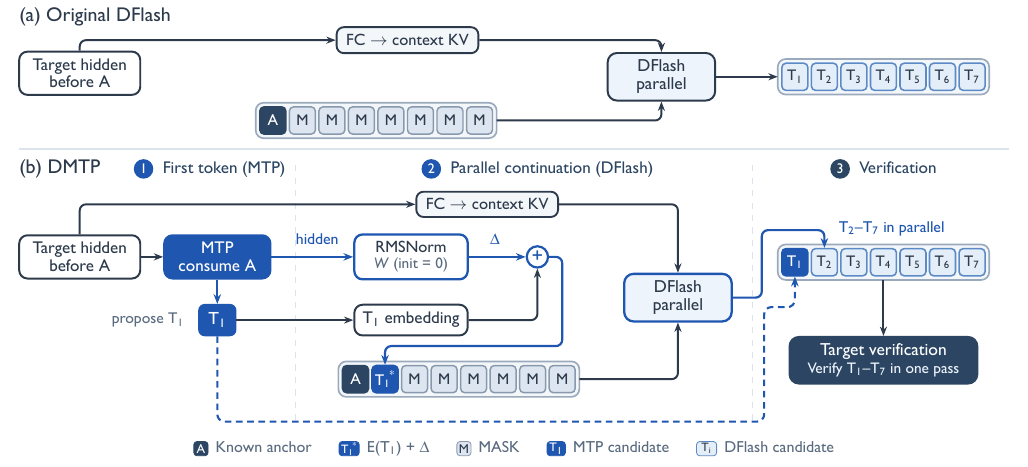}
\caption{\textbf{Overview of original DFlash (top) and DMTP (bottom) for a seven-token draft block.} Native MTP proposes $T_1$ from the committed anchor $A$ and aligned target hidden states. A zero-initialized residual adds MTP information to the $T_1$ embedding. DFlash then uses the dual anchors $[A, T_1]$ to predict $T_2$--$T_7$ in parallel, and the target verifies all seven candidates.}
\label{fig:dmtp-architecture}
\end{figure}

Figure~\ref{fig:dmtp-architecture} illustrates a complete DMTP decoding round. DMTP splits block drafting into two cooperative stages: native MTP proposes the first token $T_1$, and DFlash predicts the remaining tokens $T_2,\ldots,T_K$ in a single parallel pass. Crucially, DMTP modifies only the block input sequence fed into DFlash, leaving both backbones, DFlash's target-context KV injection, and the target model's parallel verification unchanged. As a result, DMTP is orthogonal to and directly composable with existing MTP and DFlash optimizations.

\subsection{MTP First-Token Proposal}
Let $x_{\leq a}$ denote the committed token sequence before a drafting round, ending at the anchor token $A=x_a$. Taking $A$ and its aligned target hidden states as input, native MTP computes a predictive hidden state $h_{\mathrm{mtp}}$ and samples the first draft token $T_1$ for position $a+1$ (with subsequent draft tokens $T_i$ corresponding to positions $a+i$).

Importantly, $h_{\mathrm{mtp}}$ is the pre-logit hidden state already produced while predicting $T_1$ from $A$, rather than the output of a second MTP forward pass over $T_1$. Running native MTP for just one step ($m=1$) secures the critical first token, whose predictor is trained at full pretraining scale, without incurring the linear latency of multi-step sequential rollouts (we analyze the multi-step latency trade-off in Appendix~\ref{app:dmtp-latency} and discuss adaptive scheduling in Section~\ref{sec:dmtp-limitations}). In preliminary experiments, directly repurposing native MTP layers into a DFlash-style parallel drafter degraded first-token accuracy and converged poorly, motivating us to keep single-step MTP initiation separate from parallel continuation.

\subsection{Dual-Anchor Shifted Parallel Continuation}
Whereas original DFlash predicts all $K$ tokens at once from the committed anchor $A$, DMTP places the sampled first token $T_1$ directly into the draft input and shifts parallel prediction one step forward:
\begin{align}
\text{DFlash:}&\quad (A,\mathrm{MASK}^{K})\longrightarrow T_1,\ldots,T_K,\\
\text{DMTP:}&\quad (A,T_1,\mathrm{MASK}^{K-1})\longrightarrow T_2,\ldots,T_K.
\end{align}
Here $\mathrm{MASK}$ denotes a learnable mask token. Under this layout, $T_2$ becomes DFlash's first predicted position. In our default seven-token setup ($K=7$), MTP proposes one token and DFlash proposes six.

Keeping both $A$ and $T_1$ forms \emph{dual anchors} at the start of the parallel block. Although DFlash attends to both tokens via bidirectional attention, they play different roles during verification: $A$ is already committed from the previous round, whereas $T_1$ is a draft candidate verified alongside $T_2,\ldots,T_K$ by the target model. Conditioning on a concrete $T_1$ directly simplifies parallel continuation, allowing every mask position to attend to a chosen first token rather than averaging over competing possibilities for $T_1$.

\subsection{\texorpdfstring{$T_1$}{T1}-Only Residual Fusion}
While the token embedding $E(T_1)$ captures the discrete identity of the sampled first token, $h_{\mathrm{mtp}}$ carries richer predictive context from the MTP module. To combine both signals without disturbing pretrained embeddings, we add a residual vector exclusively at the $T_1$ position, leaving $A$ and all $\mathrm{MASK}$ embeddings untouched.

Specifically, let $E(\cdot)$ denote the shared token embedding table and $\operatorname{RMSNorm}(\cdot)$ denote RMS normalization~\citep{zhang2019rmsnorm}. We normalize the detached MTP hidden state, project it through a rank-256 bottleneck $W_{\mathrm{up}}W_{\mathrm{down}}$ with a zero-initialized up-projection $W_{\mathrm{up}}$, and add the resulting residual solely to the $T_1$ embedding to form the block input $Z$:
\begin{align}
\Delta &= W_{\mathrm{up}}W_{\mathrm{down}}\,\operatorname{RMSNorm}\!\left(\operatorname{sg}(h_{\mathrm{mtp}})\right),
\qquad W_{\mathrm{up}}\big|_{\mathrm{init}}=0,\label{eq:dmtp-residual}\\
Z &= [E(A),\ E(T_1)+\Delta,\ E(\mathrm{MASK}),\ldots,E(\mathrm{MASK})],
\label{eq:dmtp-input}
\end{align}
where $\operatorname{sg}(\cdot)$ denotes the stop-gradient operator. Restricting $\Delta$ to the $T_1$ slot lets subsequent mask positions leverage MTP cues through self-attention without adding input tokens, altering DFlash's target-context KV projection, or incurring runtime overhead. Zero-initializing $W_{\mathrm{up}}$ starts training cleanly from $E(T_1)$, while the rank-256 bottleneck constrains early residual gradients ($W_{\mathrm{down}}$ receives zero initial gradient) and avoids abrupt perturbations to pretrained draft representations.

\subsection{Serving Runtime and Overhead}
\label{sec:dmtp-runtime}
\begin{wrapfigure}{r}{0.53\linewidth}
\vspace{-15pt}
\centering
\includegraphics[width=\linewidth]{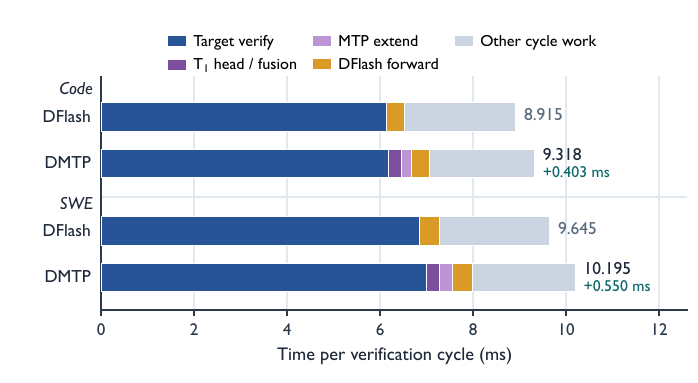}
\caption{\textbf{Per-round latency breakdown on four GPUs (TP4, batch size 8, CUDA Graphs).} Single-step MTP initiation adds only $0.403\,\mathrm{ms}$ (+4.5\%) per round on Code and $0.550\,\mathrm{ms}$ (+5.7\%) on SWE, while DFlash execution time remains virtually unchanged (exact values in Appendix~\ref{app:dmtp-latency}).}
\label{fig:dmtp-latency}
\vspace{-20pt}
\end{wrapfigure}
Adding an MTP step before DFlash improves real-world serving only if the handoff between the two modules adds minimal wall-clock latency. In our SGLang engine~\citep{sglang}, we keep the seven-token draft budget and standard verification rules, and introduce four runtime optimizations to avoid CPU-GPU stalls. Specifically, we decouple steady-state $T_1$ prediction from post-verification MTP state updates, implement DFlash's 4,096-token anchor-relative window via temporary FlashAttention-3 page tables without copying persistent KV caches, maintain a dedicated eight-slot MTP KV cache with on-device valid-row indexing, and fuse $T_1$ LM-head projection, candidate writeback, embedding lookup, and residual addition into a single CUDA Graph.

Figure~\ref{fig:dmtp-latency} profiles per-round latency on eight identical Code and SWE prompts at batch size 8 (see Appendix~\ref{app:dmtp-systems} for full system details). DFlash graph replay remains nearly identical ($0.386$ vs.\ $0.388\,\mathrm{ms}$ on Code, and $0.442$ vs.\ $0.427\,\mathrm{ms}$ on SWE), and whole-round latency increases by only $0.403\,\mathrm{ms}$ (+4.5\%) on Code and $0.550\,\mathrm{ms}$ (+5.7\%) on SWE. This small overhead is split roughly evenly between the single MTP decoder layer ($0.214$--$0.283\,\mathrm{ms}$) and $T_1$ vocabulary projection, embedding, and fusion ($0.274$--$0.290\,\mathrm{ms}$). Section~\ref{sec:dmtp-serving-results} shows how this lightweight initiation translates into end-to-end multi-GPU serving speedups.

\section{Training Recipes}
\label{sec:dmtp-training}

Training DMTP mirrors its inference workflow: native MTP provides the first-token signal, while the parallel drafter learns to predict the remaining tokens conditioned on it. Below, we first describe our default frozen-MTP recipe, used both when training from scratch and when adapting existing checkpoints, and then introduce a joint MTP-draft fine-tuning recipe for specialized domains.

\subsection{Frozen-MTP Continuation Training}
Given a training sequence, the frozen target model extracts context features up to anchor $A=x_a$, and the frozen MTP module consumes $A$ and its aligned target hidden state to compute $h_{\mathrm{mtp}}$. Under teacher forcing, the parallel drafter receives the ground-truth first token $x_{a+1}$ at the $T_1$ slot along with the detached state $h_{\mathrm{mtp}}$, and predicts the remaining tokens in parallel. In our primary seven-token DFlash setup ($K=7$), we optimize position-decayed cross-entropy over the six continuation positions:
\begin{equation}
\mathcal L_{\mathrm{tail}}=-\sum_{i=2}^{7}\alpha_i\log q_{\theta,i}(x_{a+i}\mid x_{\leq a},x_{a+1}),
\label{eq:dmtp-ce}
\end{equation}
where $\alpha_i$ are the position-decay weights from DFlash~\citep{chen2026dflash,dpace}, and no draft loss is placed on the $T_1$ input slot. During offline SFT, we use the ground-truth token $x_{a+1}$ as $T_1$, whereas inference uses the token proposed by MTP. This is an instance of the off-policy distribution mismatch inherent in offline SFT: DFlash likewise trains on recorded prefixes but encounters generated prefixes at inference. Under left-to-right verification, continuation tokens can be committed only after $T_1$ is accepted. Throughout frozen-MTP training, the target model, shared embeddings, LM head, and native MTP remain fixed. Gradients update only the parallel drafter, the $T_1$ residual adapter, and, when paired with DFlash2~\citep{dflash2}, its native block-local mixing layers and path selector under DFlash2's adaptive continuation loss.

\subsection{From-Scratch Training and Rapid Checkpoint Adaptation}
Because DMTP couples the two modules strictly through the $T_1$ token and its zero-initialized residual without altering either backbone or the target-context KV interface, it is compatible with a wide range of MTP and DFlash architectures (such as standard DFlash, DFlash2 with block-local mixing and path selection, and other parallel block drafters or first-token predictors) as well as their training objectives.

\paragraph{Training from scratch.}
When training a parallel drafter from random initialization, we freeze the target model and native MTP and optimize the draft backbone, residual adapter, and optional selector directly with the shifted continuation objective above. In our controlled Qwen3.6-27B comparison against DFlash2 (Section~\ref{sec:dmtp-scratch}), both methods use the same parallel draft backbone, matched training hyperparameters, and identical training data, with Agent trajectories intentionally excluded to evaluate out-of-distribution generalization.

\paragraph{Adapting off-the-shelf checkpoints.}
Because Equations~\ref{eq:dmtp-residual}--\ref{eq:dmtp-input} leave the target-context KV projection unchanged and initialize $\Delta=0$ at $E(T_1)$, DMTP can also be initialized directly from off-the-shelf DFlash or DFlash2 checkpoints instead of trained from scratch. Keeping the target model and MTP frozen, we fine-tune only the draft layers, selector, and residual adapter on SFT data. In our checkpoint-adaptation comparisons, DMTP and the DFlash2 baseline are fine-tuned on the same 128K-context SFT data for the same small number of updates. DMTP adapts to the shifted dual-anchor input and outperforms the matched-budget baseline across workloads (Section~\ref{sec:dmtp-ckpt}).

\subsection{Joint MTP-Draft Adaptation for Specialized Domains}
\label{sec:dmtp-joint}
On specialized downstream domains, a frozen general-purpose MTP head can fall behind a domain-tuned parallel drafter on the first token, limiting overall block acceptance. For such settings, we introduce a joint SFT recipe that adapts MTP's first-token predictor and DFlash2's parallel continuation together, while keeping the target backbone, shared token embeddings, and LM head frozen.

Specifically, in our eight-token ($K=8$) joint setup, let $p_i$ denote the frozen target distribution at block position $i$, and let $q_i$ denote the MTP distribution for $i=1$ and DFlash2's base-head distribution for $i\geq 2$. As a surrogate for improving rejection-sampling acceptance~\citep{bebop,lklosses}, we minimize the log distribution-overlap (LogTV) loss:
\begin{equation}
\ell_{\mathrm{LogTV}}(p,q)=-\log\left(\max\left\{\sum_{v}\min(p(v),q(v)),\,10^{-6}\right\}\right).
\label{eq:dmtp-logtv}
\end{equation}
When DFlash2 includes a path selector on continuation positions ($i\geq 2$), we supervise it with forward KL divergence over the candidate support $S_i$ obtained from the top-16 logits of DFlash2's base head at each position, denoting the renormalized target and selector distributions on $S_i$ by $\bar p_i^{S_i}$ and $s_i^{S_i}$. The selector scores these candidates using predecessor-token information. At inference, continuation proposals are sampled from the resulting conditional selector distribution, rather than directly from the full-vocabulary base-head distribution $q_i$ used in LogTV. The per-block joint objective is
\begin{equation}
\mathcal L=\sum_{i=1}^{8} w_i\ell_{\mathrm{LogTV}}(p_i,q_i)
+\sum_{i=2}^{8} v_i\operatorname{KL}\!\left(\bar p_i^{S_i}\Vert s_i^{S_i}\right),
\label{eq:dmtp-joint}
\end{equation}
with exponentially decayed weights $w_i\propto\exp(-(i-1)/4)$ normalized over $i\in\{1,\ldots,8\}$ and $v_i\propto\exp(-(i-2)/4)$ normalized over $i\in\{2,\ldots,8\}$ (along with standard padding masks and MoE auxiliary losses where applicable). Crucially, we keep $h_{\mathrm{mtp}}$ detached ($\operatorname{sg}(h_{\mathrm{mtp}})$) at the residual interface: the $i=1$ LogTV term supervises MTP, while continuation and selector losses do not backpropagate into MTP. This separates the first-token and continuation objectives without assuming that gradient separation alone improves accuracy.

\section{Experiments}
\label{sec:dmtp-experiments}

We evaluate DMTP along three dimensions: (1) accepted length across long-context reasoning and public short-context benchmarks, (2) end-to-end serving throughput on four GPUs, and (3) architectural ablations and training strategies (Section~\ref{sec:dmtp-extensions}).

\subsection{Experimental Setup}
\label{sec:dmtp-protocol}
\paragraph{Metrics.}
Throughout all evaluations, we aggregate speculative decoding metrics at the verification-round level (per-verify), giving equal weight to every verification step. Let $H_j$ count verification rounds that accept exactly $j$ draft tokens. We report the mean accepted length as $1+\sum_j jH_j/\sum_j H_j$, which includes the target model's correction or bonus token (differing slightly from emitted tokens per round only at sequence termination). Position-wise curves report cumulative acceptance $\Pr(N\geq i)$, the fraction of verification rounds in which all of the first $i$ draft tokens are accepted, excluding the bonus token. Workload names (e.g., Code, Agent, SWE~\citep{swebench}, and Terminal Bench~\citep{terminalbenchpaper}) denote the prompt and trajectory distributions used to measure decoding efficiency.

\paragraph{Verification and sampling protocols.}
Our long-context Qwen3.6-35B-A3B benchmark (Section~\ref{sec:dmtp-main-results}), original-OPB Qwen3.6-27B evaluation (Section~\ref{sec:dmtp-public}), and four-GPU serving benchmark (Section~\ref{sec:dmtp-serving-results}) use target-only sampling with a draft block size of $K=7$ for DFlash, DFlash2, and DMTP. The regenerated-OPB Qwen3.6-27B benchmark, Qwen3.8-Next checkpoint adaptation, and joint fine-tuning experiments (Sections~\ref{sec:dmtp-scratch}--\ref{sec:dmtp-joint-results}) use exact rejection sampling. For sequential Native MTP in Table~\ref{tab:dmtp-main}, we adopt the standard 4/1/5 configuration: four sequential drafting steps with top-1 selection, verifying five tokens including the anchor ($K=4$ draft tokens). In our serving stack, three-step MTP has drafting latency comparable to five-layer DFlash. Since DMTP adds one MTP step before DFlash, four-step MTP provides an approximately cost-matched sequential baseline (Appendix~\ref{app:dmtp-latency}) and matches our production setup.

\paragraph{Model and data settings.}
We evaluate DMTP across three target models spanning 27B to 120B parameters: Qwen3.6-27B, Qwen3.6-35B-A3B, and Qwen3.8-Next. Table~\ref{tab:dmtp-main} groups all methods by target model and training setting so that comparisons are strictly controlled within each block (MT-Bench in Panel~(a) is a separate long-context rollout set). To ensure fair comparisons, DFlash, DFlash2, and DMTP within every block share the exact same target model, parallel draft backbone, training dataset, and training budget, whether trained from scratch or fine-tuned from an existing checkpoint. In all cases, Native MTP directly uses the pretrained MTP module shipped natively with each target checkpoint, without additional draft pretraining, and stays frozen in both DMTP and the Native MTP baseline except in our joint domain fine-tuning study (Section~\ref{sec:dmtp-joint-results}). The two Qwen3.6-27B blocks use distinct setups: original OpenPerfectBlend (OPB)~\citep{openperfectblend} evaluated without thinking mode, and regenerated OPB (3K training sequence limit, three epochs) evaluated with thinking mode enabled. Long-context benchmarks use a 128K input limit and a 64K output limit, while short-context benchmarks use a 4K input limit, a 1,024-token output limit, TP4, batch limit 32, natural EOS termination, and standard two-turn MT-Bench. The default sampling temperature is $1.0$.

\begin{table}[!t]
\centering
\caption{\textbf{Main speculative decoding results across target models and evaluation settings.} All entries report per-verify accepted length including the correction or bonus token (see Section~\ref{sec:dmtp-protocol} for full details). \textbf{Bold} marks the best result within each block.}
\label{tab:dmtp-main}
\label{tab:dmtp-public}
\label{tab:dmtp-scratch}
\label{tab:dmtp-ckpt}
\begingroup
\definecolor{tblbanner}{HTML}{F4F6F8}
\definecolor{dmtprow}{HTML}{EAF2FC}
\small
\setlength{\tabcolsep}{4pt}
\renewcommand{\arraystretch}{1.24}
\newcolumntype{R}{>{\raggedleft\arraybackslash}X}
\textbf{(a) Workload long-context benchmarks}\par\smallskip
\begin{tabularx}{\linewidth}{@{}>{\columncolor{white}[0pt][\tabcolsep]}l R R r R R r R R >{\columncolor{white}[\tabcolsep][0pt]}R@{}}
\toprule
Drafter & Code & Math & MT-Bench & Agent & SWE & Long-horizon & General & Reason & Search \\
\midrule
\rowcolor{tblbanner}\multicolumn{10}{c}{\textit{Qwen3.6-35B-A3B (from scratch, target-only sampling)}} \\
Native MTP & 3.4854 & 3.6054 & 3.0320 & 4.1695 & 3.2618 & 3.9074 & 3.1931 & 3.5239 & 3.0843 \\
DFlash & 3.2159 & 3.4440 & 2.7852 & 4.5266 & 2.8113 & 3.9424 & 2.6962 & 3.2907 & 2.6909 \\
\rowcolor{dmtprow}DMTP & \textbf{3.8217} & \textbf{4.0500} & \textbf{3.2445} & \textbf{5.0986} & \textbf{3.6053} & \textbf{4.3635} & \textbf{3.2722} & \textbf{3.8761} & \textbf{3.2916} \\
\midrule
\rowcolor{tblbanner}\multicolumn{10}{c}{\textit{Qwen3.6-27B (OPB-regen, thinking, from scratch, rejection sampling)}} \\
Native MTP & 3.5697 & 3.6913 & 3.1467 & 4.1280 & 3.2583 & 3.7297 & 3.2227 & 3.5974 & 3.1624 \\
DFlash2 & 3.9621 & 4.2909 & 4.0044 & 3.8197 & 2.9273 & 3.1929 & 3.3412 & 4.1282 & 3.4653 \\
\rowcolor{dmtprow}DMTP & \textbf{4.6022} & \textbf{4.8871} & \textbf{4.1826} & \textbf{4.6871} & \textbf{3.9427} & \textbf{4.3310} & \textbf{3.7736} & \textbf{4.7106} & \textbf{4.0437} \\
\midrule
\rowcolor{tblbanner}\multicolumn{10}{c}{\textit{Qwen3.8-Next (from checkpoint, rejection sampling)}} \\
Native MTP & 3.0136 & 3.0916 & 2.9210 & 3.7712 & 3.7817 & 3.5106 & \textbf{2.8158} & 2.9740 & \textbf{3.2090} \\
DFlash2 & 2.8793 & 2.9534 & 2.8380 & 3.8963 & 4.0989 & 3.5280 & 2.5339 & 2.7902 & 2.9158 \\
\rowcolor{dmtprow}DMTP & \textbf{3.2858} & \textbf{3.3600} & \textbf{2.9879} & \textbf{4.2819} & \textbf{4.4628} & \textbf{4.0195} & 2.7944 & \textbf{3.2578} & 3.1454 \\
\bottomrule
\end{tabularx}
\par\medskip
\textbf{(b) Public short-context benchmarks}\par\smallskip
\begin{tabularx}{\linewidth}{@{}>{\columncolor{white}[0pt][\tabcolsep]}l R R R R >{\columncolor{white}[\tabcolsep][0pt]}R@{}}
\toprule
Drafter & GSM8K & MATH500 & HumanEval & MT-Bench & Alpaca \\
\midrule
\rowcolor{tblbanner}\multicolumn{6}{c}{\textit{Qwen3.6-27B (original OPB, non-thinking, from scratch, target-only sampling)}} \\
Native MTP & 4.373 & 4.377 & 4.698 & 3.599 & 3.529 \\
DFlash & 4.318 & 4.258 & 6.424 & 3.209 & 2.763 \\
\rowcolor{dmtprow}DMTP(D) & 5.074 & 4.968 & 6.795 & 3.760 & 3.358 \\
DFlash2 & 4.595 & 4.536 & 6.609 & 3.535 & 2.990 \\
\rowcolor{dmtprow}DMTP(D2) & \textbf{5.371} & \textbf{5.233} & \textbf{7.041} & \textbf{4.052} & \textbf{3.643} \\
\midrule
\rowcolor{tblbanner}\multicolumn{6}{c}{\textit{Qwen3.6-27B (OPB-regen, thinking, from scratch, rejection sampling)}} \\
Native MTP & 3.9737 & 3.9968 & 3.6617 & 3.2994 & 3.2405 \\
DFlash2 & 5.8619 & 5.7963 & 5.1567 & 4.4048 & 4.2018 \\
\rowcolor{dmtprow}DMTP & \textbf{5.9383} & \textbf{5.9702} & \textbf{5.3305} & \textbf{4.4654} & \textbf{4.3257} \\
\bottomrule
\end{tabularx}
\endgroup
\end{table}

\subsection{Long-Context Reasoning and Agentic Acceptance}
\label{sec:dmtp-main-results}
We first train DFlash and DMTP from scratch for three epochs on the same 32K-context SFT data with matched training hyperparameters on Qwen3.6-35B-A3B~\citep{yang2025qwen3}. The training set contains roughly 250K sequences of up to 32K tokens, drawn mainly from agentic and long-horizon trajectories, while MT-Bench~\citep{mtbench} serves as an out-of-domain chat testbed. Both methods are evaluated on identical prompts at temperature $1.0$ under the length limits in Section~\ref{sec:dmtp-protocol}. Code, Math, and MT-Bench pair short prompts with long reasoning outputs, whereas Agent, SWE, and Long-horizon pair long input contexts with shorter action responses.

As shown in the 35B block of Table~\ref{tab:dmtp-main}(a), DMTP improves per-verify accepted length over DFlash by 10.68\%--28.25\% across all nine workloads. It also surpasses four-step Native MTP on every task, including Agent and Long-horizon where a seven-token parallel block reaches further than four sequential steps. These results support our core design: using native MTP for $T_1$ removes DFlash's first-token bottleneck, while parallel continuation extends the accepted prefix without multi-step MTP latency.

\begin{figure}[!t]
\centering
\includegraphics[width=\linewidth]{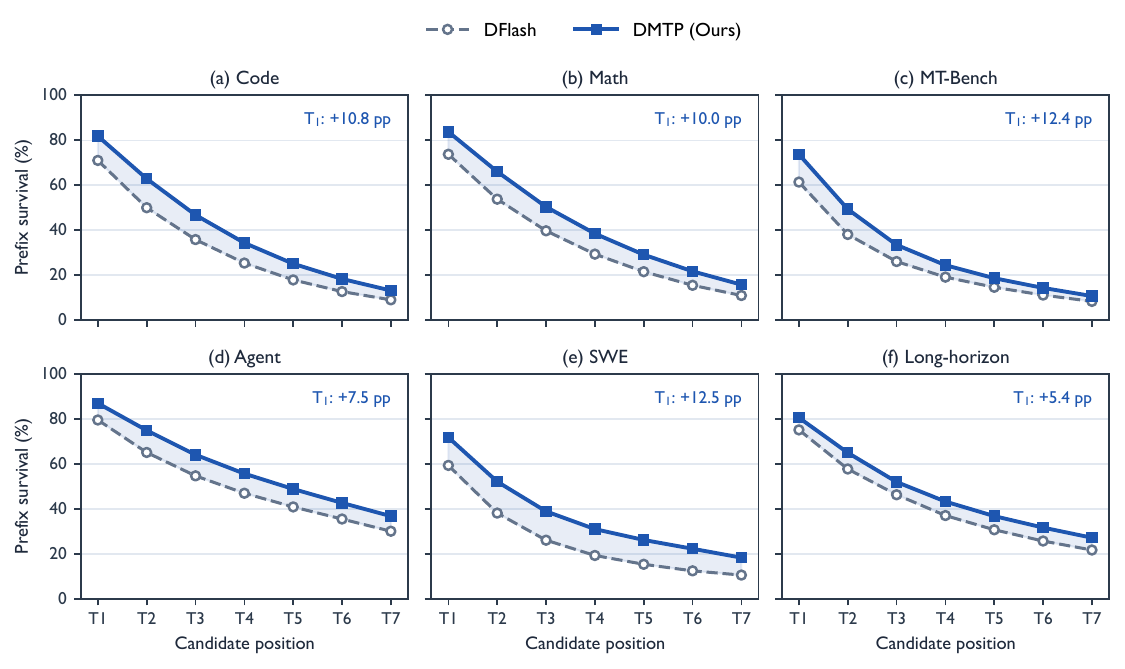}
\caption{\textbf{Cumulative prefix acceptance across the seven draft positions on Qwen3.6-35B-A3B.} Each point shows the fraction of verification rounds accepting all candidates up to position $i$ (excluding bonus tokens). DMTP lifts acceptance at $T_1$ and maintains a clear lead across $T_2$--$T_7$.}
\label{fig:dmtp-main-position}
\end{figure}

Figure~\ref{fig:dmtp-main-position} breaks down these gains across the seven draft positions. During closed-loop rollouts, DMTP raises first-token acceptance over DFlash by 5.43--12.46 percentage points, for example, from 71.00\% to 81.81\% on Code, 73.73\% to 83.72\% on Math, and 59.33\% to 71.79\% on SWE (minor differences from Figure~\ref{fig:dmtp-first-token} come from different prefixes visited during closed-loop generation). More importantly, this margin holds across the entire block: pairing a stronger $T_1$ with $T_1$-conditioned continuation improves cumulative acceptance from $T_2$ all the way to $T_7$.

\subsection{Public Short-Context Benchmarks}
\label{sec:dmtp-public}
To evaluate DMTP in a fully reproducible public setting, we train five-layer drafters on the public Qwen3.6-27B model using OpenPerfectBlend, a math- and code-heavy SFT corpus. All methods share the same training setup: dynamic packing up to 2K tokens, global batch size 128 on eight GPUs in BF16, and three epochs, with the target model and native MTP kept frozen. DFlash uses position-decayed cross-entropy, while DFlash2 adds block-local mixing, path selection, and an adaptive continuation loss. DMTP(D) and DMTP(D2) equip DFlash and DFlash2 with our dual-anchor $T_1$-residual interface under the same $K=7$ budget. On the original-OPB split, we evaluate 80 prompts per benchmark with \texttt{enable\_thinking=False} (preserving each model's own first-turn response on two-turn MT-Bench).

In the original-OPB block of Table~\ref{tab:dmtp-public}(b), DMTP(D) improves accepted length over DFlash by 5.78\%--21.55\% across all five benchmarks (GSM8K, MATH500~\citep{math500}, HumanEval~\citep{humaneval}, MT-Bench~\citep{mtbench}, and Alpaca), already surpassing standalone DFlash2~\citep{dflash2} on every dataset. Combining DMTP with DFlash2 yields further gains: DMTP(D2) improves over DFlash2 by 6.54\%--21.84\% (reaching up to 7.041 on HumanEval and 5.371 on GSM8K) and outperforms both DFlash2 and Native MTP across the board. This confirms that DMTP is orthogonal to DFlash2's block-internal refinements and composes cleanly with them. Notably, the largest relative gains appear on out-of-domain chat benchmarks (+21.84\% on Alpaca and +14.63\% on MT-Bench for DMTP(D2)), where native MTP bridges the domain gap left by math- and code-heavy draft SFT. Correspondingly, per-verify $T_1$ acceptance for DMTP(D) rises by 5.79--16.59 percentage points over DFlash across all five benchmarks, jumping from 66.86\% to 83.45\% on Alpaca and from 67.27\% to 83.23\% on MT-Bench (full breakdown in Appendix Table~\ref{tab:dmtp-public-t1}).

Panel~(b) also reports regenerated-OPB checkpoints evaluated with thinking enabled under exact rejection sampling (64 conversations per benchmark, temperature $1.0$, and a 1,024-token output cap per turn, so it is not directly comparable to the non-thinking block above). Across all five benchmarks, DMTP outperforms both DFlash2 and Native MTP, exceeding four-step Native MTP by 33.49\%--49.44\% (e.g., 5.9383 vs.\ 3.9737 on GSM8K and 5.9702 vs.\ 3.9968 on MATH500). Because regenerated OPB closely matches short-context math and code, DFlash2 is already strong in-domain and the short-context margin over DFlash2 is modest. However, since draft training excludes Agent and long-context data, reusing native MTP produces much wider gains on held-out Agent, SWE, and Long-horizon workloads in Table~\ref{tab:dmtp-main}(a) (Section~\ref{sec:dmtp-scratch}).

\subsection{End-to-End Serving Throughput}
\label{sec:dmtp-serving-results}
Longer accepted prefixes speed up real serving only if the MTP step remains lightweight under batched execution. We therefore measure end-to-end throughput on Qwen3.6-35B-A3B across four GPUs (TP4) in SGLang~\citep{sglang} under active-batch limits $B\in\{2,4,8,16,32\}$. All runs use temperature $1.0$, natural EOS termination, a 28K output limit, and a 262,144-token server context pool, flushing KV caches between runs while keeping standard intra-run RadixAttention prefix reuse. At the primary B32 setting, DFlash and DMTP run back-to-back on the same node over 64 identical requests per workload.

\begin{figure}[!t]
\centering
        \includegraphics[width=\linewidth]{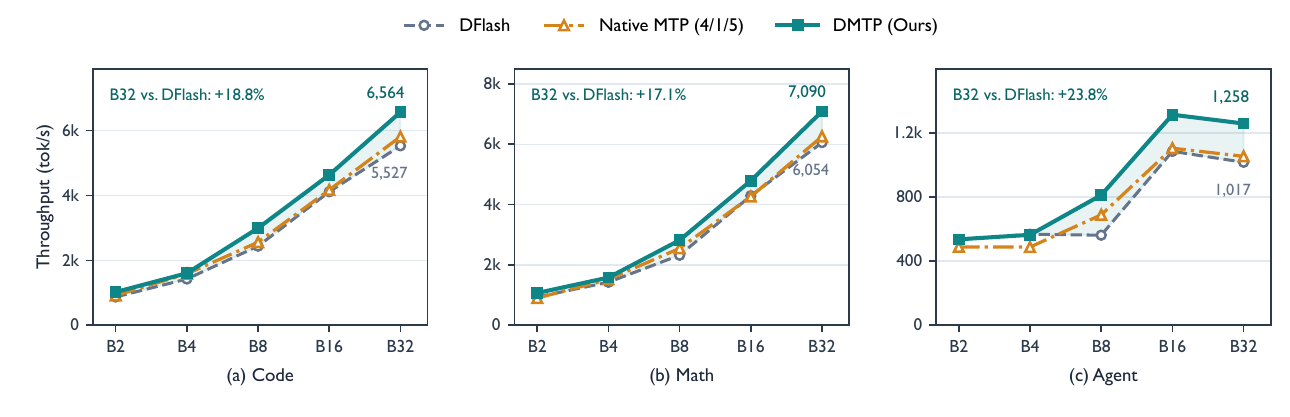}
\caption{\textbf{End-to-end serving throughput across active-batch limits on four GPUs (Qwen3.6-35B-A3B, TP4).} Throughput includes request queueing, prompt prefill, drafting, and target verification under target-only sampling across $B\in\{2,4,8,16,32\}$, alongside a Native MTP (4/1/5) reference sweep (full values and setup in Appendix~\ref{app:dmtp-batch-sweep}). At B32, DMTP improves throughput over DFlash by 18.77\% on Code (5,527 to 6,564\,tok/s), 17.10\% on Math (6,054 to 7,090\,tok/s), and 23.75\% on long-context Agent (1,017 to 1,258\,tok/s).}
\label{fig:dmtp-serving}
\end{figure}

As shown in Figure~\ref{fig:dmtp-serving}, DMTP outperforms both DFlash and the four-step Native MTP (4/1/5) reference at every batch limit on Code and Math, and achieves substantial speedups on long-context Agent from B8 to B32. At B32, DMTP raises end-to-end throughput over DFlash by 17.10\%--23.75\% across all three workloads, while also beating Native MTP (4/1/5) by 12.97\% on Code, 13.28\% on Math, and 19.42\% on Agent. On Agent at very small batch limits ($B\le 4$), prefill latency, short outputs, and scheduling overhead partly offset the longer accepted prefixes (Appendix~\ref{app:dmtp-batch-sweep}). From B8 onward, KV-heavy target verification dominates round time. The additional per-round latency of 0.40--0.55\,ms (Figure~\ref{fig:dmtp-latency}, Appendix~\ref{app:dmtp-latency}) is readily amortized, turning longer accepted prefixes directly into higher serving throughput.

\subsection{Architectural Ablations and Training Strategies}
\label{sec:dmtp-extensions}

Having established DMTP's offline acceptance and serving throughput, we next ablate its hidden-state interface and anchor layout, and evaluate three practical training settings under exact rejection sampling: from-scratch training with DFlash2, fast adaptation of off-the-shelf DFlash2 checkpoints, and joint MTP-draft fine-tuning on specialized domains.

\subsubsection{Where Should MTP Hidden Representations Enter?}
\label{sec:dmtp-hidden-analysis}
To test how MTP's hidden state $h_{\mathrm{mtp}}$ best complements the discrete token $T_1$, we compare six injection interfaces on an eight-request Code benchmark at $K=7$ (Figure~\ref{fig:dmtp-ablation-design}a). Aside from original DFlash (which anchors on $A$), all variants here use single-anchor $T_1$ conditioning so that only the hidden-state interface varies.

Conditioning on the discrete token $T_1$ alone raises accepted length from 3.3348 to 3.7574. Injecting $h_{\mathrm{mtp}}$ at the $T_1$ position improves acceptance further, reaching 3.8492 (+2.44\%) with zero-initialized residual addition and 3.9480 (+5.07\%) with full token-hidden fusion, and outperforms broadcasting the residual across all six mask positions (+1.98\%). By contrast, fusing $h_{\mathrm{mtp}}$ into or replacing the multi-layer target context hurts acceptance (-1.65\% and -5.60\%), showing that MTP's local next-token state complements rather than replaces multi-layer target KV features. Although full token-hidden fusion reaches a slightly higher score when trained from scratch with a single anchor, our lightweight $T_1$ residual enriches $E(T_1)$ with extra MTP hidden information without overwriting the token embedding itself, pairing much more consistently with dual-anchor inputs below.

\subsubsection{Single versus Dual Anchors}
\label{sec:dmtp-anchor-analysis}
\begin{figure}[!t]
\centering
\includegraphics[width=\linewidth]{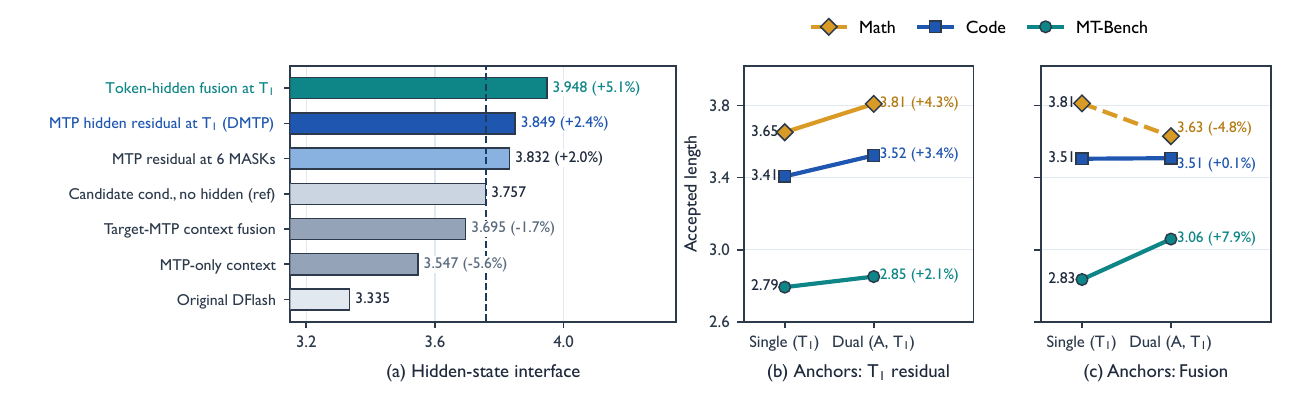}
\caption{\textbf{Architectural ablations on hidden-state interfaces and anchor layouts.} (a)~Per-verify accepted length across hidden-state interfaces on eight Code requests ($K=7$). The dashed line at 3.7574 shows discrete $T_1$ conditioning without MTP hidden states, and the six-MASK residual bar comes from an auxiliary run on identical inputs. (b,c)~Single-anchor versus dual-anchor layouts: keeping the committed anchor $A$ alongside $T_1$ consistently improves accepted length under $T_1$ residual injection (b), whereas full token-hidden fusion shows mixed task sensitivity (c).}
\label{fig:dmtp-ablation-design}
\end{figure}

We next compare a single-anchor input $(T_1,\allowbreak\mathrm{MASK}^{K-1})$ against our dual-anchor input $(A,\allowbreak T_1,\allowbreak\mathrm{MASK}^{K-1})$, which keeps the committed token $A$ alongside the proposed token $T_1$. Figure~\ref{fig:dmtp-ablation-design}(b,c) evaluates both layouts under either $T_1$ residual injection or full token-hidden fusion across Code, Math, and MT-Bench.

With $T_1$ residual injection, keeping $A$ consistently improves accepted length across all three domains, whereas full token-hidden fusion behaves inconsistently across tasks. Pairing dual anchors with $T_1$-only residual injection exposes the local token transition $(A, T_1)$ directly to DFlash while preserving a clean token-embedding interface.

\subsubsection{From-Scratch Compatibility with DFlash2 and OOD Generalization}
\label{sec:dmtp-scratch}
To verify that DMTP also strengthens next-generation parallel drafters under exact rejection sampling, we train DFlash2 and dual-anchor DMTP from scratch on Qwen3.6-27B using regenerated OpenPerfectBlend sequences up to 3K tokens, keeping the target model and native MTP frozen. Both drafters share identical training configurations and data, and are evaluated with thinking mode enabled. Crucially, the training data contains no Agent trajectories, providing a clean test of out-of-distribution generalization.

Both DFlash2 and DMTP are trained for three epochs on the same regenerated OpenPerfectBlend data. As shown in the regenerated-27B block of Table~\ref{tab:dmtp-scratch}(a), dual-anchor DMTP outperforms both DFlash2 (+4.45\%--35.64\%) and four-step Native MTP (+13.54\%--32.92\%) on every workload. The largest gains over DFlash2 appear on the held-out Agent (+22.71\%), SWE (+34.69\%), and Long-horizon (+35.64\%) suites, where standalone Native MTP already beats DFlash2. Because native MTP was pretrained on broad corpora alongside the target model, passing $T_1$ and $h_{\mathrm{mtp}}$ into DFlash2 gives the parallel drafter a reliable starting point even on domains unseen during draft SFT.

\subsubsection{Rapid Adaptation of Off-the-Shelf Checkpoints}
\label{sec:dmtp-ckpt}
When strong DFlash2~\citep{dflash2} checkpoints already exist, DMTP does not need to be trained from scratch. Following the frozen-MTP adaptation recipe in Section~\ref{sec:dmtp-training}, we initialize DMTP directly from an open-source DFlash2 checkpoint on Qwen3.8-Next~\citep{qsa} and fine-tune on 128K-context SFT data. For a matched comparison, the DFlash2 baseline is also fine-tuned on the same data for the same small number of updates. Both models are evaluated under exact rejection sampling at batch size 32, temperature $1.0$, 128K max input length, and 64K max output length.

As shown in the Qwen3.8-Next block of Table~\ref{tab:dmtp-ckpt}(a), matched-budget fine-tuning improves accepted length over DFlash2 across all nine workloads. Adapted DMTP also surpasses four-step Native MTP on seven of nine tasks, including Code, Math, MT-Bench, and Reason, where DFlash2 alone trails Native MTP. It nearly matches Native MTP on General (2.7944 vs.\ 2.8158) and Search (3.1454 vs.\ 3.2090). First-token acceptance rises across all long-context suites: from 61.80\% to 67.40\% on General, 69.46\% to 77.04\% on Reason, 76.82\% to 82.10\% on Agent, 74.03\% to 81.43\% on Long-horizon, 81.96\% to 86.86\% on SWE, and 63.23\% to 67.62\% on Search.

We also test checkpoint adaptation when the starting parallel drafter already matches native MTP on the first token, fine-tuning a continued-pretrained DFlash2 checkpoint ($K=8$). Both DMTP and its DFlash2 baseline are fine-tuned on the same 128K-context SFT data for the same small number of updates. Even when DFlash2 starts with comparable or slightly higher $T_1$ acceptance than native MTP, shifted continuation lets DMTP beat DFlash2 at every continuation position ($T_2$--$T_8$) across all six workloads (42 of 42 entries), and even surpass sequential multi-step MTP on most continuation slots (see Appendix~\ref{app:dmtp-cpt} for the full breakdown). This further confirms that prefix shifting brings structural gains independent of first-token accuracy.

\subsubsection{Joint MTP-Draft Adaptation on Specialized Tasks}
\label{sec:dmtp-joint-results}
Finally, we evaluate joint fine-tuning (Section~\ref{sec:dmtp-joint}) on a specialized task where a frozen general-purpose MTP head becomes the bottleneck. Starting from the adapted checkpoint above with the target model frozen, we run offline SFT on collected Terminal Bench~\citep{terminalbenchpaper} rollout traces, comparing frozen-MTP adaptation against joint MTP-draft adaptation under exact rejection sampling ($K=8$, concurrency 32, 255 evaluation requests).

\begin{figure}[!t]
\centering
\includegraphics[width=\linewidth]{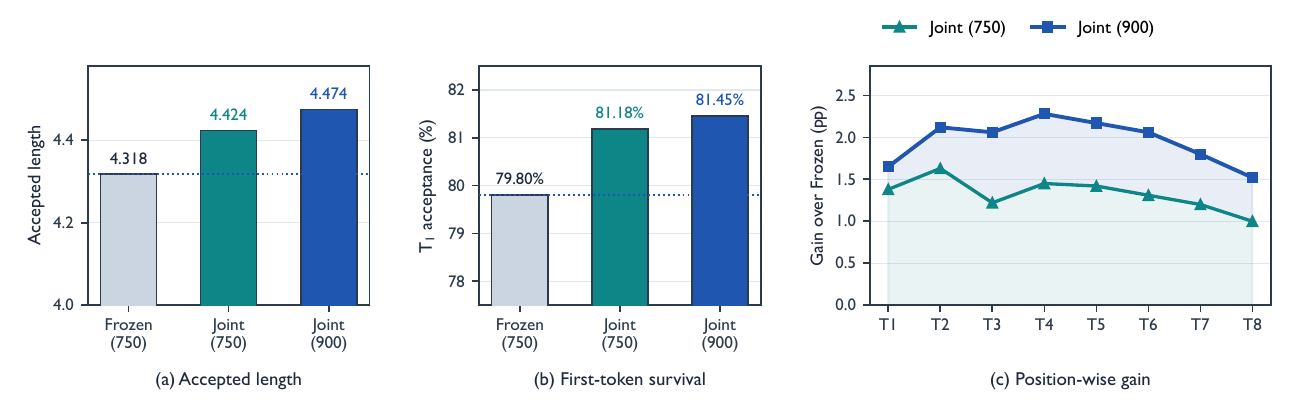}
\caption{\textbf{Joint MTP-draft SFT on Terminal Bench traces ($K=8$, rejection sampling).} At a matched training budget, jointly fine-tuning MTP and DFlash2 improves both per-verify accepted length (a, +2.46\%) and first-token acceptance (b, +1.38\,pp), lifting cumulative acceptance at every position $T_1$--$T_8$ (c), with further gains from extended joint training.}
\label{fig:dmtp-joint-summary}
\end{figure}

At a matched training budget (Figure~\ref{fig:dmtp-joint-summary}), joint fine-tuning raises first-token acceptance by 1.38 percentage points (from 79.80\% to 81.18\%) and accepted length by 2.46\%, improving cumulative acceptance across all eight positions $T_1$--$T_8$ (Appendix~\ref{app:dmtp-joint-figure}). Extending joint training reaches an accepted length of 4.4745 (+3.63\%). Joint adaptation and prefix shifting thus address complementary bottlenecks: joint training sharpens first-token prediction on specialized domains, while prefix shifting strengthens parallel continuation across the rest of the block.

\section{Related Work}
\label{sec:dmtp-related}
\paragraph{Parallel and semi-autoregressive block drafting.}
Parallel drafters reduce sequential drafting latency by predicting multiple tokens in one backbone pass. Falcon uses semi-autoregressive decoding trees~\citep{falcon}, PARD adapts small models to parallel mask prediction~\citep{pard}, and P-EAGLE turns EAGLE into a parallel predictor~\citep{peagle}. DFlash~\citep{chen2026dflash} conditions a bidirectional block on multi-layer target KV features, while DFlare~\citep{dflare} uses layer-wise feature fusion. Recent work primarily improves intra-block dependencies and later-position accuracy: Domino~\citep{domino} adds a causal refinement head, DSpark~\citep{cheng2026dspark} pairs a sequential head with load-aware verification, and DFlash2~\citep{dflash2} combines local convolutions, path selection, and adaptive continuation losses. DMTP addresses a complementary issue: first-token accuracy and OOD generalization. It lets DFlash reuse MTP's pretrained capabilities and conditions parallel continuation on the sampled $T_1$, gaining from prefix shifting even when first-token accuracies are similar. This change precedes the parallel backbone and is orthogonal to block-internal refinements. We demonstrate compatibility with DFlash and DFlash2, while combining it with other causal heads remains a natural extension.

\paragraph{Speculative decoding, draft trees, and feature-level drafters.}
Speculative decoding speeds up LLM inference by verifying lightweight draft tokens in parallel with the target model, preserving the exact target distribution under rejection sampling~\citep{leviathan2023fast,chen2023accelerating}. Subsequent work reduces drafting cost through auxiliary decoding heads or layer skipping~\citep{medusa,zhang2024draft,swift,knnssd}, improves candidate coverage via tree verification and dynamic pruning~\citep{specinfer,li2024eagle2,hu2026echo,liu2026talon,liu2026dcut}, or reuses candidates from retrieval, cached transitions, target logits, and overlapped execution~\citep{rest,tokenrecycling,liu2025logitspec,shen2026graft,pearl,specbranch,double}. Feature-level drafters predict from target hidden states: EAGLE~\citep{li2024eagle,li2025eagle3} predicts future features autoregressively, KVShot~\citep{liu2026kvshot} reuses target KV caches, and native MTP~\citep{gloeckle2024bmtp} learns future-token prediction jointly during pretraining, as adopted in DeepSeek-V3~\citep{deepseekv3} and Qwen3.5~\citep{qwen35}. Broad pretraining gives MTP strong first-token accuracy and domain coverage, but a $K$-step rollout incurs $O(K)$ sequential latency. DMTP combines these strengths with DFlash's low-cost parallel drafting: it runs MTP only once, then delegates the remaining block to DFlash rather than extending the sequential rollout.

\paragraph{Draft training, acceptance objectives, and RL serving.}
Recent work also improves draft accuracy through knowledge distillation~\citep{distillspec}, on-policy distillation on draft-generated prefixes~\citep{gkd,draftopd,specopd}, and online updates from live serving traces~\citep{wang2026aurora}. Because verification stops at the first rejected token, newer training losses replace uniform cross-entropy with position-aware weighting~\citep{dpace}, expected decoding rounds~\citep{edr}, or direct distribution-overlap objectives~\citep{lklosses,bebop,specopd}. Speculative decoding is also increasingly used to accelerate reinforcement learning rollouts and long-tail serving~\citep{liu2025specrl,chang2026srt,iso2026rlspec,bebop,xu2026tailsieve,DBLP:journals/corr/abs-2610-07767,DBLP:journals/corr/abs-2607-15810,wang2026hybridcache}. DMTP builds on these acceptance-aligned losses in both frozen-MTP adaptation and joint LogTV fine-tuning, while detaching $h_{\mathrm{mtp}}$ to keep first-token and continuation gradients cleanly separated.

\section{Limitations and Future Directions}
\label{sec:dmtp-limitations}
DMTP relies on native MTP to propose the first token, whose advantage depends on target pretraining coverage and domain alignment. When a domain-specialized parallel drafter already matches or exceeds frozen MTP on $T_1$, keeping MTP frozen limits potential gains unless both modules are fine-tuned together. Several practical directions follow naturally from our study:
\begin{itemize}[leftmargin=5mm,itemsep=3pt,topsep=3pt]
    \item \textbf{Reducing $T_1$ LM-head latency.} As shown in Appendix~\ref{app:dmtp-latency}, projecting $T_1$ over the full vocabulary accounts for roughly half of DMTP's added per-round latency ($0.27$--$0.29\,\mathrm{ms}$). Lightweight drafting heads, such as low-rank shortlisting with exact rescoring as in EAGLE-4~\citep{eagle4} or draft-side LM-head quantization, could cut this cost further without changing target verification.
    \item \textbf{Adaptive multi-step MTP and unified models.} Instead of fixing one MTP step per round, an adaptive scheduler could run extra MTP steps when token uncertainty is high or serving load is light. On the modeling side, unifying MTP and parallel drafting or training multi-layer MTP modules directly for block prediction could clarify how architecture, hidden-state transfer, and data scale each shape first-token versus continuation accuracy.
    \item \textbf{On-policy and online joint training.} Our frozen-MTP recipe uses ground-truth $x_{a+1}$ at $T_1$, and our joint LogTV recipe trains offline on collected traces. Exploring on-policy draft sampling~\citep{gkd,draftopd,specopd}, balancing joint losses without hurting OOD generalization, and integrating online RL updates~\citep{wang2026aurora,bebop} are promising next steps.
    \item \textbf{Sparse verification and load-aware scheduling.} Extending our serving evaluation from target-only sampling to sparse top-$k$ rejection sampling (Appendix~\ref{app:dmtp-sparse-rs}), sparse block verification, and load-aware dynamic block sizing~\citep{cheng2026dspark,liu2026dcut} merits further system profiling.
    \item \textbf{Extending to target models without native MTP.} Appendix~\ref{app:dmtp-eagle-first-token} shows that with enough training data, a single-step EAGLE predictor reaches first-token accuracy close to native MTP. Replacing native MTP with an EAGLE-style initiator~\citep{li2024eagle,li2025eagle3} and jointly training EAGLE with DFlash could bring prefix-shifted parallel drafting to any target model, including those pretrained without native MTP.
\end{itemize}

\section{Conclusion}
We presented DMTP, a cooperative speculative decoding framework that pairs native Multi-Token Prediction with parallel block drafting. In each round, native MTP proposes the critical first token and injects extra predictive hidden information into DFlash via a lightweight residual on $T_1$. DFlash then conditions on dual anchors to predict the remaining tokens in parallel. This design combines MTP's strong first-token accuracy and OOD generalization with prefix-shifted parallel continuation, making subsequent tokens easier to predict without paying multi-step sequential latency. Across short- and long-context benchmarks on Qwen3.6-27B, Qwen3.6-35B-A3B, and Qwen3.8-Next, DMTP consistently outperforms DFlash and DFlash2, surpasses four-step Native MTP on nearly all workloads, and delivers 17.10\%--23.75\% higher end-to-end serving throughput on four GPUs. Because DMTP operates at the interface between MTP and parallel drafting, it is orthogonal to and directly composable with existing MTP and DFlash optimizations, supporting scratch training, checkpoint adaptation, and joint domain fine-tuning as a simple, effective, and deployment-ready path for fast LLM inference.

\section*{Acknowledgements}
This work was supported by Alibaba Research Intern Program. We thank the SpecForge team, especially Jiaping Wang, for their support in adapting the training code for DMTP.

\FloatBarrier
\bibliographystyle{plainnat}
\bibliography{dmtp_reference}

\clearpage
\appendix

\pdfbookmark[0]{Appendix Contents}{dmtp-appendix-contents}
\section*{Appendix Contents}
\begingroup
\hypersetup{linkcolor=black}
\small
\newcommand{\appendixtocsection}[2]{\par\addvspace{6pt}\noindent\textbf{\hyperref[#1]{\makebox[3.2em][l]{\ref*{#1}}#2}}\nobreak\dotfill\hyperref[#1]{\pageref*{#1}}\par}
\newcommand{\appendixtocsubsection}[2]{\noindent\hspace*{1.5em}\hyperref[#1]{\makebox[3.2em][l]{\ref*{#1}}#2}\nobreak\dotfill\hyperref[#1]{\pageref*{#1}}\par}
\appendixtocsection{app:dmtp-systems}{Inference System Design and Latency Breakdown}
\appendixtocsubsection{app:dmtp-sys-opts}{Runtime Optimizations and State Management}
\appendixtocsubsection{app:dmtp-latency}{Measured Per-Round Latency Breakdown and Cost Structure}
\appendixtocsubsection{app:dmtp-sparse-rs}{Sparse Rejection Sampling}
\appendixtocsection{app:dmtp-supplementary}{Supplementary Experiments and Analysis}
\appendixtocsubsection{app:dmtp-exp-summary}{Experimental and Model Configuration Summary}
\appendixtocsubsection{app:dmtp-release-residual}{Residual Adapters in the Released Checkpoints}
\appendixtocsubsection{app:dmtp-capacity-budget}{Parameter Budget and Draft Depth}
\appendixtocsubsection{app:dmtp-metric-details}{Verification-Round Metrics, Per-Request Aggregation, and Prefix Survival}
\appendixtocsubsection{app:dmtp-conditional}{Conditional Acceptance Across the Draft Block}
\appendixtocsubsection{app:dmtp-shift-values}{Complete Progressive Shift Ablation and Cross-Model First-Token Behavior}
\appendixtocsubsection{app:dmtp-batch-sweep}{Batch-Limit Sensitivity of End-to-End Serving}
\appendixtocsubsection{app:dmtp-public-first-token}{First-Candidate Acceptance on Public Short-Context Benchmarks}
\appendixtocsubsection{app:dmtp-7k-first-token}{First-Token Survival After Checkpoint Adaptation}
\appendixtocsubsection{app:dmtp-eagle-first-token}{First-Token Predictors Trained at Scale}
\appendixtocsection{app:dmtp-figure-data}{Numerical Data Underlying the Figures}
\appendixtocsubsection{app:dmtp-figure-first}{First-Token and Cumulative Acceptance}
\appendixtocsubsection{app:dmtp-figure-interfaces}{Hidden-State Interfaces and Anchor Layouts}
\appendixtocsubsection{app:dmtp-figure-joint}{Terminal Bench Joint-Training Curves}
\appendixtocsection{app:dmtp-cpt}{Position-Wise Survival Under Matched First-Token Accuracy}
\appendixtocsection{app:dmtp-resources}{Open-Source Artifacts}
\appendixtocsection{app:dmtp-joint-figure}{Extended Joint MTP-Draft Training Analysis}
\appendixtocsection{app:dmtp-related-detailed}{Detailed Related Work}
\appendixtocsubsection{app:dmtp-related-exact}{Exact Speculative Sampling and Prefix Survival}
\appendixtocsubsection{app:dmtp-related-parallel}{Parallel and Block-Diffusion Drafters}
\appendixtocsubsection{app:dmtp-related-causal}{Intra-Block Causality and Semi-Autoregressive Drafting}
\appendixtocsubsection{app:dmtp-related-mtp}{Native MTP and Feature-Conditioned Drafting}
\appendixtocsubsection{app:dmtp-related-training}{Draft Distillation and Acceptance-Aligned Objectives}
\appendixtocsubsection{app:dmtp-related-serving}{Verification Scheduling and Serving Efficiency}
\appendixtocsubsection{app:dmtp-related-rl}{RL Rollouts and Joint Domain Adaptation}
\appendixtocsubsection{app:dmtp-related-multimodal}{Multimodal Speculative Decoding and Cache Compression}
\endgroup
\clearpage

\section{Inference System Design and Latency Breakdown}
\label{app:dmtp-systems}
This section describes the SGLang~\citep{sglang} serving implementation behind our throughput benchmarks (Section~\ref{sec:dmtp-serving-results}), explains how we keep the handoff between native MTP and DFlash fast under CUDA Graphs, and presents the full per-round latency breakdown for Figure~\ref{fig:dmtp-latency}.

\subsection{Runtime Optimizations and State Management}
\label{app:dmtp-sys-opts}
\begin{figure}[htbp]
\centering
\includegraphics[width=\linewidth]{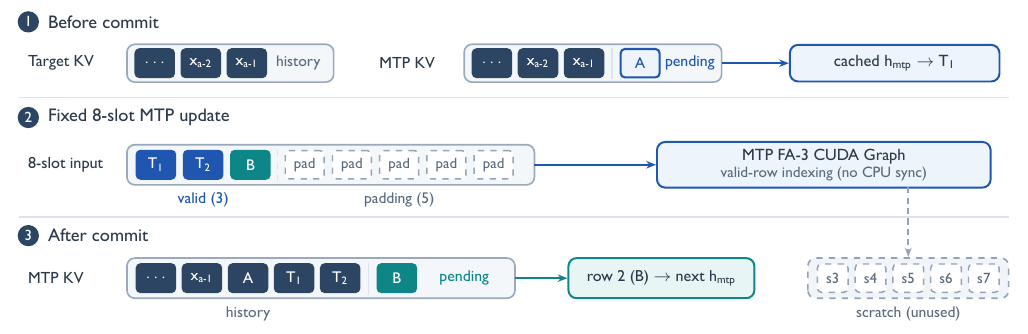}
\caption{\textbf{Dedicated MTP KV cache lifecycle ($T_1, T_2$ accepted, token $B$ emitted).} A fixed eight-slot CUDA Graph updates MTP without CPU syncs, committing $A, T_1, T_2$ to history, placing $B$ in the pending slot for the next $h_{\mathrm{mtp}}$, and routing unused slots to scratch space.}
\label{fig:dmtp-kv-lifecycle}
\end{figure}

Keeping the seven-token draft budget and standard target verification unchanged, our serving engine focuses on two newly added stages: post-verification MTP state updates and steady-state $T_1$ preparation. Figures~\ref{fig:dmtp-runtime-flow} and~\ref{fig:dmtp-kv-lifecycle} show the steady-state execution flow and dedicated MTP KV cache lifecycle. Four coordinated optimizations remove CPU-GPU synchronization and keep both drafters inside static CUDA Graphs (Table~\ref{tab:dmtp-sys-opts}):

\begin{figure}[htbp]
\centering
\includegraphics[width=\linewidth]{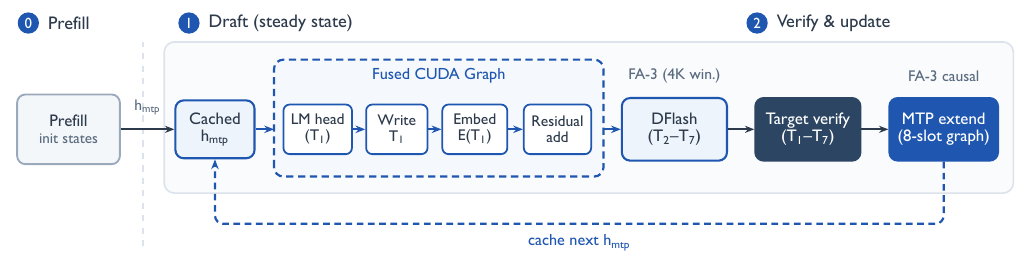}
\caption{\textbf{Decoupled steady-state execution flow in SGLang.} Each round starts from the cached $h_{\mathrm{mtp}}$, runs $T_1$ preparation in one fused CUDA Graph, drafts $T_2$--$T_7$ in parallel, verifies $T_1$--$T_7$, and advances MTP to cache the next $h_{\mathrm{mtp}}$.}
\label{fig:dmtp-runtime-flow}
\end{figure}

\begin{itemize}[leftmargin=5mm,itemsep=4pt,topsep=4pt]
    \item \textbf{Decoupling prediction from state updates.} Prompt prefill initializes the target model, MTP, and DFlash together and caches the initial MTP hidden state $h_{\mathrm{mtp}}$. Each steady-state round therefore starts directly from the cached $h_{\mathrm{mtp}}$: the runtime projects $T_1$ through the LM head, looks up its embedding, adds the MTP residual, and runs one parallel DFlash pass. Only after target verification commits the accepted prefix and emits the correction or bonus token does the engine advance MTP over the newly committed tokens, caching the next $h_{\mathrm{mtp}}$ so the following round again begins immediately at the $T_1$ head.
    \item \textbf{Unified FlashAttention-3 backend with separate attention masks.} Both MTP and DFlash run on FlashAttention-3 inside CUDA Graphs while respecting their own attention rules. Native MTP uses causal attention over the full committed history, whereas DFlash attends to the 4,096 context tokens before the anchor along with its intra-block mask. For DFlash, temporary page tables define both the anchor-relative context window and the block mask without copying persistent KV caches or changing rotary position (RoPE) indices, ensuring every draft query attends to the exact anchor-aligned window rather than a shifted per-query window. For MTP, FlashAttention-3 runs directly on fixed eight-slot on-device metadata, avoiding CPU-to-GPU length transfers during normal decoding.
    \item \textbf{Dedicated MTP KV cache lifecycle (history, pending, and scratch).} Because MTP can consume the current anchor slightly ahead of target commit, we give MTP a dedicated KV cache split into committed \emph{history}, a \emph{pending} anchor slot, and an eight-slot per-request on-device \emph{scratch} buffer. Using a fixed eight-slot update buffer lets the same pre-captured CUDA Graph replay no matter how many draft tokens were accepted. At commit time, an on-device valid-row counter routes accepted tokens into history, places the bonus or correction token into the next pending slot, and selects the corresponding hidden state as the next $h_{\mathrm{mtp}}$, while unused trailing slots write safely to scratch memory and are ignored in later steps.
    \item \textbf{Exact-batch CUDA Graphs and fused $T_1$ preparation.} We capture CUDA Graphs for exact active batch sizes so fusion projections run only over active requests without padding waste. All metadata updates happen on GPU before graph replay. To remove host launch gaps between small kernels, we merge $T_1$ LM-head projection, candidate writeback, target embedding lookup, and residual addition into a single CUDA Graph, preserving exact numerical behavior, argmax selection, and tensor-parallel communication.
\end{itemize}

\begin{table}[htbp]
\centering
\caption{\textbf{Summary of SGLang inference optimizations for DMTP.} On-device metadata, fixed eight-slot update buffers, temporary FlashAttention-3 page tables, and combined CUDA Graphs keep the handoff between native MTP and DFlash lightweight.}
\label{tab:dmtp-sys-opts}
\begingroup
\footnotesize
\setlength{\tabcolsep}{6pt}
\renewcommand{\arraystretch}{1.20}
\newcolumntype{L}{>{\raggedright\arraybackslash}X}
\begin{tabularx}{\linewidth}{@{}l L L@{}}
\toprule
Component & Adaptation / optimization & Eliminated overhead \\
\midrule
MTP advancement & Device-side fixed 8 slots, dedicated KV cache, graph replay & Per-step CPU length transfers, pack/expand, repeated launches \\
DFlash attention & Temporary FlashAttention-3 page tables for anchor window and block mask & Rebuilding or copying persistent KV caches across windows \\
$T_1$ preparation & Combined LM head $\to$ candidate write $\to$ embedding $\to$ fusion graph & Multiple host graph launches and inter-stage syncs \\
State commit & Valid-row indexing into history, pending, and scratch slots & Dynamic-length memory layout conversions \\
\bottomrule
\end{tabularx}
\endgroup
\end{table}

\FloatBarrier
\subsection{Measured Per-Round Latency Breakdown and Cost Structure}
\label{app:dmtp-latency}
Once host launch and layout overheads are removed, DMTP's extra compute per round consists of one native MTP decoder layer and the full-vocabulary LM head used to sample $T_1$. Because the MTP layer runs full-history causal attention along with linear projections, normalization, feed-forward layers, and tensor-parallel all-reduces, its latency grows slightly with context length. Meanwhile, the LM head reads the full vocabulary projection matrix to compute logits over all tokens even though only one token $T_1$ is sampled.

To measure each stage on production hardware, we profile standalone DFlash against DMTP on four GPUs (TP4) with identical target weights. Both Code and SWE run eight identical prompts at an active batch size of 8 with CUDA Graphs enabled, generating 512 tokens with EOS ignored after 16 warmup rounds. Table~\ref{tab:dmtp-latency-breakdown} lists the stream-level latency means plotted in Figure~\ref{fig:dmtp-latency}.

\begin{table}[htbp]
\centering
\caption{\textbf{Measured per-round latency breakdown (ms/round) on four GPUs (TP4, batch size 8, CUDA Graphs).} DFlash graph replay stays nearly unchanged on both workloads, while whole-round latency rises by $0.403\,\mathrm{ms}$ (+4.52\%) on Code and $0.550\,\mathrm{ms}$ (+5.70\%) on SWE.}
\label{tab:dmtp-latency-breakdown}
\begingroup
\definecolor{dmtprow}{HTML}{EAF2FC}
\footnotesize
\setlength{\tabcolsep}{6pt}
\renewcommand{\arraystretch}{1.18}
\newcolumntype{R}{>{\raggedleft\arraybackslash}X}
\begin{tabularx}{\linewidth}{@{}l R R R R@{}}
\toprule
Stage (ms / round) & Code: DFlash & Code: DMTP & SWE: DFlash & SWE: DMTP \\
\midrule
DFlash graph replay & 0.386 & 0.388 & 0.442 & 0.427 \\
Additional MTP graph replay & -- & 0.214 & -- & 0.283 \\
Additional $T_1$ head, embedding, and fusion & -- & 0.290 & -- & 0.274 \\
\midrule
\rowcolor{dmtprow}\textbf{Whole round (incl.\ prep.\ and verification)} & \textbf{8.915} & \textbf{9.318} & \textbf{9.645} & \textbf{10.195} \\
\bottomrule
\end{tabularx}
\endgroup
\end{table}

Across both workloads, DFlash graph replay stays nearly identical ($0.386$ vs.\ $0.388\,\mathrm{ms}$ on Code, and $0.442$ vs.\ $0.427\,\mathrm{ms}$ on SWE). Whole-round latency increases by just $0.403\,\mathrm{ms}$ (+4.52\%) on Code and $0.550\,\mathrm{ms}$ (+5.70\%) on SWE. The additional stages comprise the MTP decoder layer ($0.214$--$0.283\,\mathrm{ms}$, slightly higher on longer SWE contexts) and $T_1$ projection, embedding, and fusion ($0.274$--$0.290\,\mathrm{ms}$). Because the full-vocabulary LM head accounts for about half of this extra cost, lightweight drafting heads, such as EAGLE-4's low-rank shortlisting with exact rescoring or draft LM-head quantization, offer a straightforward way to trim initiation latency further without changing target verification. This profile also motivates our 4/1/5 Native MTP baseline in Table~\ref{tab:dmtp-main} and Figure~\ref{fig:dmtp-serving}: in our serving engine, three-step MTP (MTP314) has drafting latency similar to five-layer DFlash, so adding one MTP step in DMTP matches the drafting budget of four-step MTP (MTP415), which is also the standard Qwen serving baseline.

More generally, if a cooperative drafter runs $m$ sequential MTP steps (each taking $T_{\mathrm{mtp}}$ for the MTP layer and LM head) followed by one parallel DFlash pass ($T_{\mathrm{dflash}}$) and target verification ($T_{\mathrm{verify}}$), advancing by $L(m)$ tokens per round on average, the amortized per-token latency is
\begin{equation}
\frac{T_{\mathrm{verify}} + m T_{\mathrm{mtp}} + T_{\mathrm{dflash}}}{L(m)}.
\label{eq:dmtp-m-tradeoff}
\end{equation}
We use $m=1$ to balance acceptance and drafting latency: (1) only the first MTP step is trained across the full pretraining corpus while governing the critical first-token rate $c_1$ (Equation~\ref{eq:dmtp-supp-tau}), and (2) each extra sequential step ($m \ge 2$) adds $T_{\mathrm{mtp}} \approx 0.50$--$0.56\,\mathrm{ms}$ of serial latency while bringing diminishing gains $L(m)-L(m-1)$ once first-token uncertainty is resolved. Adaptive policies that increase $m$ selectively on uncertain tokens remain an interesting direction (Section~\ref{sec:dmtp-limitations}).

\FloatBarrier
\subsection{Sparse Rejection Sampling}
\label{app:dmtp-sparse-rs}
In addition to target-only sampling and dense rejection sampling, our released SGLang engine supports \emph{sparse rejection sampling} across all draft positions, including $T_1$. Instead of computing and all-gathering full-vocabulary draft distributions across tensor-parallel ranks, the drafter keeps only the top-$k$ candidate logits and token IDs at each position, greatly cutting TP communication and memory traffic. To preserve the exact target distribution, the saved sparse draft probabilities match the renormalized top-$k$ distribution actually used during draft sampling, assigning zero draft probability outside the top-$k$ set. Combining sparse rejection sampling with sparse block verification and profiling end-to-end serving speedups is left to future work.
\FloatBarrier

\section{Supplementary Experiments and Analysis}
\label{app:dmtp-supplementary}
This appendix summarizes all model and training configurations, defines our aggregation metrics, and reports position-wise conditional acceptance rates, progressive shift ablations, batch-limit serving sweeps, and supplementary first-token studies supporting Section~\ref{sec:dmtp-experiments}. Comparisons are grouped by target model and evaluation setting. Candidate budgets are specified separately: Native MTP uses four draft tokens in the main comparisons, while DFlash and DMTP use seven; the extended position-wise study uses eight.

\subsection{Experimental and Model Configuration Summary}
\label{app:dmtp-exp-summary}
Because we evaluate DMTP across several target models, training setups, and verification protocols, Table~\ref{tab:dmtp-exp-summary} collects the key settings from Sections~\ref{sec:dmtp-motivation}--\ref{sec:dmtp-extensions} in one place for easy reference.

\begin{table}[!htbp]
\centering
\caption{\textbf{Summary of target models, draft configurations, training regimes, and evaluation protocols.}}
\label{tab:dmtp-exp-summary}
\begingroup
\footnotesize
\setlength{\tabcolsep}{4.5pt}
\renewcommand{\arraystretch}{1.18}
\newcolumntype{L}{>{\raggedright\arraybackslash}X}
\begin{tabularx}{\linewidth}{@{}l L l c L L@{}}
\toprule
Target model & Role / study & Draft backbone & $K$ & Verification / eval & Training regime \\
\midrule
Qwen3.6-35B-A3B & Main long-context \& TP4 serving & DFlash (5-layer) & 7 & Target-only; $T{=}1.0$; 128K in / 64K out (28K serving) & Scratch; $\sim$250K seqs (32K, Agent/Long-horizon) \\
Qwen3.6-27B & Public short-context (orig.\ OPB) & DFlash, DFlash2 (5-layer) & 7 & Target-only; $T{=}1.0$; non-think; 1K out & Scratch; OPB (2K pack, BS 128, 3 epochs) \\
Qwen3.6-27B & Long-context OOD \& short think & DFlash2 & 7 & Exact RS; $T{=}1.0$; thinking; 128K/64K \& 1K & Scratch; OPB-regen (3K limit, 3 epochs, no Agent) \\
Qwen3.7-Turbo & Progressive shift ablation (\S\ref{sec:dmtp-ablation}) & DFlash & 7 & Target-only; Code 128K, LH 80K, MT-Bench & Scratch; matched-training checkpoints (1--2\,pp MTP $T_1$ gap) \\
Qwen3.8-Next & Checkpoint adaptation (\S\ref{sec:dmtp-ckpt}, \ref{app:dmtp-cpt}) & DFlash2 (open / CPT) & 7 / 8 & Exact RS; $T{=}1.0$; BS 32; 128K in / 64K out & From ckpt (frozen MTP); SFT ($K{=}7$) / adaptation ($K{=}8$) \\
\bottomrule
\end{tabularx}
\endgroup
\end{table}

\subsection{Residual Adapters in the Released Checkpoints}
\label{app:dmtp-release-residual}
Both released checkpoints inject MTP features exclusively at the $T_1$ embedding via a zero-initialized residual branch, differing only in the internal parameterization of the adapter. Let $e=E(T_1)$ and $h=\operatorname{sg}(h_{\mathrm{mtp}})$. The released \href{https://huggingface.co/Piggy12/Qwen3.6-27B-DMTP}{Qwen3.6-27B checkpoint} implements the normalized-hidden residual interface in Section~\ref{sec:dmtp-method} with a hidden-only rank-256 bottleneck:
\begin{equation}
\Delta_{27\mathrm{B}}=W_{\mathrm{up}}\operatorname{SiLU}\!\left(W_{\mathrm{down}}\operatorname{RMSNorm}(h)\right),
\qquad W_{\mathrm{up}}\big|_{\mathrm{init}}=0.
\end{equation}
The released \href{https://huggingface.co/Piggy12/Qwen3.8-next-120b-dmtp}{Qwen3.8-Next-120B checkpoint} instead normalizes the token embedding and MTP hidden state separately, concatenates them, and applies a linear projection within the residual branch:
\begin{equation}
\Delta_{120\mathrm{B}}=W_{\mathrm{cat}}\left[\operatorname{RMSNorm}_e(e);\,\operatorname{RMSNorm}_h(h)\right],
\qquad W_{\mathrm{cat}}\big|_{\mathrm{init}}=0.
\end{equation}
In both checkpoints, the parallel drafter receives $[E(A),\,e+\Delta,\,E(\mathrm{MASK}),\ldots]$, leaving the anchor embedding, all mask embeddings, and the target-context KV interface unchanged. In the 120B checkpoint, concatenation occurs strictly inside the residual branch rather than replacing $E(T_1)$ or adding an extra token slot, so zero-initializing $W_{\mathrm{cat}}$ preserves exact checkpoint compatibility at the start of adaptation.

In either case, the core mechanism is identical: conditioning DFlash on a realized $T_1$ while passing normalized MTP features through a zero-initialized $T_1$-only residual.

\subsection{Parameter Budget and Draft Depth}
\label{app:dmtp-capacity-budget}
Because DMTP pairs an unchanged parallel draft backbone with the target model's native MTP module and a lightweight residual adapter, matching the parallel backbone alone does not equalize total drafting parameters. A natural question is whether allocating a comparable parameter budget by deepening either DFlash or native MTP would yield similar gains.

\paragraph{Deepening DFlash.}
Prior depth ablations in DFlash~\citep{chen2026dflash} directly evaluate this alternative. Expanding DFlash from five to eight layers raises accepted length slightly, but fails to improve end-to-end speedup: reported speedups shift from $4.71\times$ to $4.64\times$ on MATH500, remain $3.96\times$ on HumanEval, and drop from $2.35\times$ to $2.23\times$ on MT-Bench. Simply adding parallel layers therefore yields diminishing returns in the accuracy-latency trade-off. By contrast, DMTP changes the conditioning structure itself: proposing $T_1$ beforehand shifts the parallel window forward so that every remaining token conditions on a concrete first token rather than deeper unconditioned layers.

\paragraph{Deepening MTP.}
Adding layers to native MTP can improve per-step prediction quality, but every extra layer sits directly on the sequential critical path and executes at every step of a multi-step rollout, compounding serial drafting latency. DMTP instead invokes native MTP only once for the highest-leverage first token and delegates the remaining positions to a single parallel pass. Parameter count alone therefore does not govern serving efficiency. Practical comparisons must weigh accepted length alongside per-round drafting latency (Appendix~\ref{app:dmtp-latency}).

\subsection{Verification-Round Metrics, Per-Request Aggregation, and Prefix Survival}
\label{app:dmtp-metric-details}
Let $N\in\{0,\ldots,K\}$ be the number of draft tokens accepted before the first rejection in a verification round, excluding the target model's correction or bonus token. Given a pooled histogram $H_j$ counting rounds that accept exactly $j$ draft tokens, we define cumulative acceptance $s_i=\Pr(N\geq i)$ and conditional position-wise acceptance $c_i=\Pr(N\geq i\mid N\geq i-1)$ as
\begin{equation}
s_i=\frac{\sum_{j=i}^{K}H_j}{\sum_{j=0}^{K}H_j},\qquad
c_i=\frac{s_i}{s_{i-1}},\qquad s_0=1.
\label{eq:dmtp-supp-survival}
\end{equation}
The per-verify accepted length is then
\begin{equation}
\tau = 1 + \mathbb{E}[N] = 1 + \frac{\sum_{j=0}^{K} j H_j}{\sum_{j=0}^{K} H_j} = 1 + \sum_{i=1}^{K} s_i = 1 + \sum_{i=1}^{K}\prod_{j=1}^{i} c_j,
\label{eq:dmtp-supp-tau}
\end{equation}
where the leading $1$ counts the target model's correction or bonus token (differing slightly from emitted tokens per step only when generation hits EOS or the output cap). Because $c_1$ multiplies every term in Equation~\ref{eq:dmtp-supp-tau}, improving first-token accuracy boosts cumulative acceptance across the entire block. Conversely, later conditional rates $c_i$ need not rise at every single position for $s_i$ and $\tau$ to increase. Moreover, because speculative decoding runs closed-loop rollouts, conditional rates at later positions reflect the prefixes that actually survived earlier steps rather than a fixed set of offline prompts.

In serving benchmarks, accepted length can be averaged either \emph{per request} (computing each request's mean first, then averaging equally across requests) or \emph{per verify} (Equation~\ref{eq:dmtp-supp-tau}, weighting every verification round equally). Table~\ref{tab:dmtp-per-req-vs-verify} compares both metrics on our B32 serving benchmark on Qwen3.6-35B-A3B. On Code and Math, where output lengths are fairly uniform, per-request and per-verify numbers match closely. On Agent, SWE, and Long-horizon, where output lengths vary widely, short easy responses inflate unweighted per-request averages, whereas longer, harder responses account for most verification rounds and dominate wall-clock time. We therefore report per-verify accepted length by default, as it directly reflects total verification steps and serving throughput.

\begin{table}[!htbp]
\centering
\caption{\textbf{Per-request versus per-verify accepted length on the B32 serving benchmark (Qwen3.6-35B-A3B, TP4, $K=7$).} Both metrics include the correction or bonus token across 64 identical requests per workload. The two metrics agree closely on Code and Math, whereas on variable-length agentic tasks, per-verify aggregation more accurately tracks total verification work.}
\label{tab:dmtp-per-req-vs-verify}
\begingroup
\definecolor{dmtprow}{HTML}{EAF2FC}
\footnotesize
\setlength{\tabcolsep}{6pt}
\renewcommand{\arraystretch}{1.18}
\newcolumntype{R}{>{\raggedleft\arraybackslash}X}
\begin{tabularx}{\linewidth}{@{}l R R R R R R@{}}
\toprule
& \multicolumn{3}{c}{Per-request accepted length} & \multicolumn{3}{c}{Per-verify accepted length} \\
\cmidrule(lr){2-4}\cmidrule(l){5-7}
Workload & DFlash & DMTP & Gain & DFlash & DMTP & Gain \\
\midrule
Code & 3.227 & \textbf{3.814} & +18.19\% & 3.198 & \textbf{3.795} & +18.67\% \\
Math & 3.461 & \textbf{4.100} & +18.46\% & 3.429 & \textbf{4.056} & +18.29\% \\
MT-Bench & 2.890 & \textbf{3.459} & +19.69\% & 2.706 & \textbf{3.250} & +20.10\% \\
Agent & 4.795 & \textbf{5.254} & +9.57\% & 4.249 & \textbf{4.709} & +10.83\% \\
SWE & 3.407 & \textbf{3.881} & +13.91\% & 2.794 & \textbf{3.173} & +13.56\% \\
Long-horizon & 4.356 & \textbf{5.022} & +15.29\% & 3.641 & \textbf{4.656} & +27.88\% \\
\bottomrule
\end{tabularx}
\endgroup
\end{table}

\subsection{Conditional Acceptance Across the Draft Block}
\label{app:dmtp-conditional}
To separate first-token improvements from per-step continuation accuracy, Table~\ref{tab:dmtp-conditional} complements the cumulative acceptance curves in Figure~\ref{fig:dmtp-main-position} with conditional acceptance rates $c_i$ computed from the same pooled histograms (64 requests per workload, temperature $1.0$, $K=7$, target-only sampling).

\begin{table}[!htbp]
\centering
\caption{\textbf{Conditional acceptance (\%) by draft position on Qwen3.6-35B-A3B ($K=7$, target-only sampling).} Each entry shows $c_i = \Pr(N \ge i \mid N \ge i-1)$ excluding correction/bonus tokens, and Rounds gives total verification steps across 64 requests.}
\label{tab:dmtp-conditional}
\begingroup
\definecolor{dmtprow}{HTML}{EAF2FC}
\footnotesize
\setlength{\tabcolsep}{4pt}
\renewcommand{\arraystretch}{1.16}
\newcolumntype{R}{>{\raggedleft\arraybackslash}X}
\begin{tabularx}{\linewidth}{@{}ll r R R R R R R R@{}}
\toprule
Workload & Method & Rounds & $T_1$ & $T_2$ & $T_3$ & $T_4$ & $T_5$ & $T_6$ & $T_7$ \\
\midrule
Code & DFlash & 523,201 & 71.00 & 70.38 & 71.64 & 70.79 & 70.43 & 70.91 & 70.97 \\
\rowcolor{dmtprow}\cellcolor{white}& DMTP & 434,780 & \textbf{81.81} & \textbf{76.83} & \textbf{74.34} & \textbf{73.43} & \textbf{73.00} & \textbf{73.02} & \textbf{71.76} \\
\midrule
Math & DFlash & 473,538 & 73.73 & 72.93 & 73.82 & 73.87 & 73.36 & 71.90 & 70.59 \\
\rowcolor{dmtprow}\cellcolor{white}& DMTP & 404,689 & \textbf{83.72} & \textbf{78.84} & \textbf{76.25} & \textbf{76.36} & \textbf{75.64} & \textbf{74.54} & \textbf{72.62} \\
\midrule
MT-Bench & DFlash & 54,479 & 61.36 & 62.04 & \textbf{68.34} & \textbf{73.36} & \textbf{76.38} & 76.11 & \textbf{74.92} \\
\rowcolor{dmtprow}\cellcolor{white}& DMTP & 41,259 & \textbf{73.72} & \textbf{66.90} & 67.78 & 73.03 & 76.24 & \textbf{76.77} & 74.74 \\
\midrule
Agent & DFlash & 22,213 & 79.53 & 81.84 & 84.00 & 85.87 & 87.05 & 86.81 & 84.74 \\
\rowcolor{dmtprow}\cellcolor{white}& DMTP & 17,371 & \textbf{86.98} & \textbf{86.12} & \textbf{85.51} & \textbf{86.91} & \textbf{87.72} & \textbf{87.34} & \textbf{86.18} \\
\midrule
SWE & DFlash & 4,795 & 59.33 & 64.25 & 68.38 & 73.92 & 79.65 & 80.98 & \textbf{84.90} \\
\rowcolor{dmtprow}\cellcolor{white}& DMTP & 3,836 & \textbf{71.79} & \textbf{72.59} & \textbf{74.64} & \textbf{79.69} & \textbf{84.44} & \textbf{85.06} & 82.20 \\
\midrule
Long-horizon & DFlash & 2,919 & 75.13 & 76.84 & \textbf{80.18} & 80.01 & 82.89 & 83.71 & 84.40 \\
\rowcolor{dmtprow}\cellcolor{white}& DMTP & 2,798 & \textbf{80.56} & \textbf{80.66} & 79.98 & \textbf{83.22} & \textbf{85.04} & \textbf{86.10} & \textbf{85.78} \\
\bottomrule
\end{tabularx}
\endgroup
\end{table}

Table~\ref{tab:dmtp-conditional} highlights two clear patterns. First, DMTP raises $T_1$ acceptance by 5.43--12.46 percentage points across all six workloads and simultaneously raises conditional acceptance at $T_2$ by 3.82--8.34 percentage points (since $T_2$ is now DFlash's first predicted position, conditioned on a known $T_1$). Together, these gains cut total verification rounds by 15\%--24\% on Code, Math, MT-Bench, Agent, and SWE. Second, at deeper positions ($T_3$--$T_7$ on MT-Bench and $T_7$ on SWE), DMTP's conditional rate occasionally matches or dips slightly below DFlash even though its cumulative acceptance stays much higher (Table~\ref{tab:dmtp-cumulative-source}). This is a standard selection effect in closed-loop decoding: because DMTP lets many more moderately uncertain prefixes pass $T_1$ and $T_2$, the prefixes reaching deeper slots are slightly harder on average, yet cumulative survival remains consistently higher across the entire block.

\subsection{Complete Progressive Shift Ablation and Cross-Model First-Token Behavior}
\label{app:dmtp-shift-values}
Across model families, native MTP's first-token lead over an SFT-trained parallel drafter varies with pretraining and post-training setups. On Qwen3.5, Qwen3.6, and GLM checkpoints, native MTP holds a large first-token advantage. On newer Qwen3.8 checkpoints, however, a well-trained DFlash2 drafter nearly matches native MTP on in-domain benchmarks, likely due to changes in pretraining data mixtures. Thus, native MTP's value comes not only from higher first-token accuracy on OOD tasks, but also from \emph{prefix shifting}, which helps even when first-token accuracy is tied.

When native MTP supplies $T_1$, DFlash predicts $T_2,\ldots,T_K$ conditioned on a concrete first token rather than starting from $A$. During training, $T_2$ becomes DFlash's first unknown slot and approaches standalone DFlash's $T_1$ accuracy, while every later position moves one step closer to a known prefix. Table~\ref{tab:dmtp-shift-values} gives the exact numbers for Figure~\ref{fig:dmtp-shift-motivation} on Qwen3.7-Turbo, where native MTP leads DFlash at $T_1$ by only 1--2 percentage points.

\begin{table}[!htbp]
\centering
\caption{\textbf{Progressive ablation isolating first-token replacement, hidden-state injection, and prefix shifting on Qwen3.7-Turbo.} Per-verify accepted lengths include correction/bonus tokens. \emph{Hidden-only} and \emph{Hidden+$T_1$} use checkpoints trained for the same number of updates. Adjacent deltas ($\Delta$) show the gain over the preceding column, while \emph{Shift gain} compares \emph{Hidden+$T_1$} against \emph{Hidden-only}.}
\label{tab:dmtp-shift-values}
\begingroup
\definecolor{dmtprow}{HTML}{EAF2FC}
\footnotesize
\setlength{\tabcolsep}{4.5pt}
\renewcommand{\arraystretch}{1.18}
\newcolumntype{R}{>{\raggedleft\arraybackslash}X}
\begin{tabularx}{\linewidth}{@{}l R R R R R@{}}
\toprule
Workload & DFlash & +MTP first ($\Delta$) & Hidden-only ($\Delta$) & Hidden+$T_1$ ($\Delta$) & Shift gain \\
\midrule
Code (128K) & 4.2951 & 4.4234 (+0.1283) & 4.4612 (+0.0378) & \textbf{4.7274} (+0.2662) & \textbf{+5.97\%} \\
Long-horizon (80K) & 3.7745 & 3.7638 ($-$0.0107) & 3.8341 (+0.0703) & \textbf{4.1998} (+0.3657) & \textbf{+9.54\%} \\
MT-Bench Think & 2.8207 & 2.9594 (+0.1387) & 2.9893 (+0.0299) & \textbf{3.2853} (+0.2960) & \textbf{+9.90\%} \\
MT-Bench Non-think & 2.7691 & 2.8948 (+0.1257) & 2.9111 (+0.0163) & \textbf{3.1711} (+0.2600) & \textbf{+8.93\%} \\
\bottomrule
\end{tabularx}
\endgroup
\end{table}

Because first-token accuracy is nearly tied here, simply replacing $T_1$ brings small gains ($+0.1257$ to $+0.1387$) on three tasks and a slight drop ($-0.0107$) on Long-horizon, where unshifted continuation tokens clash with the externally sampled $T_1$. Adding MTP hidden states without $T_1$ input conditioning adds only $0.0163$--$0.0703$ tokens per round. By contrast, feeding $T_1$ into DFlash and shifting continuation to $T_2$ produces the largest jump on every workload ($+0.2600$ to $+0.3657$ tokens, or $+5.97\%$ to $+9.90\%$ over \emph{Hidden-only}), confirming that prefix shifting brings substantial gains on its own.

\subsection{Batch-Limit Sensitivity of End-to-End Serving}
\label{app:dmtp-batch-sweep}
To study serving performance across concurrency levels, we sweep the active-batch limit $B\in\{2,4,8,16,32\}$ on four GPUs (TP4) under dynamic batching, temperature $1.0$, natural EOS termination, a 28K output cap, and a 262,144-token server context pool. Here $B$ caps the number of concurrently active requests as sequences arrive and finish. KV caches are flushed between runs while preserving standard intra-run RadixAttention prefix reuse. Throughput equals total output tokens divided by wall-clock completion time for the full request batch, including queueing, prefill, drafting, and target verification. Table~\ref{tab:dmtp-batch-sweep} lists the exact accepted lengths and throughputs plotted in Figure~\ref{fig:dmtp-serving}.

\begin{table}[!htbp]
\centering
\caption{\textbf{End-to-end serving sensitivity across active-batch limits on four GPUs (Qwen3.6-35B-A3B, TP4).} $R$ is the number of requests per workload. Accepted length is per-verify (including correction/bonus tokens), and throughput is total output tokens divided by wall time (tok/s). \textbf{Bold} marks the better result between same-weight, same-runtime DFlash and DMTP. Native MTP (4/1/5) is shown alongside as a reference.}
\label{tab:dmtp-batch-sweep}
\begingroup
\definecolor{dmtprow}{HTML}{EAF2FC}
\footnotesize
\setlength{\tabcolsep}{4pt}
\renewcommand{\arraystretch}{1.15}
\newcolumntype{R}{>{\raggedleft\arraybackslash}X}
\begin{tabularx}{\linewidth}{@{}l r r R R R R R R@{}}
\toprule
& & & \multicolumn{3}{c}{Accepted length (per-verify)} & \multicolumn{3}{c}{Throughput (tok/s)} \\
\cmidrule(lr){4-6}\cmidrule(l){7-9}
Workload & $B$ & $R$ & Native MTP & DFlash & DMTP & Native MTP & DFlash & DMTP \\
\midrule
Code & 2 & 8 & 3.5310 & 3.2993 & \textbf{3.8560} & 919.21 & 871.55 & \textbf{1,022.39} \\
& 4 & 8 & 3.5023 & 3.3380 & \textbf{3.9032} & 1,583.04 & 1,424.85 & \textbf{1,602.90} \\
& 8 & 8 & 3.5174 & 3.2234 & \textbf{3.9004} & 2,558.75 & 2,437.02 & \textbf{2,991.40} \\
& 16 & 64 & 3.4745 & 3.2203 & \textbf{3.7871} & 4,169.13 & 4,111.63 & \textbf{4,631.08} \\
\rowcolor{dmtprow}\cellcolor{white}& 32 & 64 & 3.4935 & 3.1979 & \textbf{3.7948} & 5,810.31 & 5,526.78 & \textbf{6,564.08} \\
\midrule
Math & 2 & 8 & 3.6230 & 3.5564 & \textbf{4.0549} & 900.13 & 936.84 & \textbf{1,069.78} \\
& 4 & 8 & 3.6359 & 3.4702 & \textbf{4.0948} & 1,520.84 & 1,426.14 & \textbf{1,581.54} \\
& 8 & 8 & 3.6140 & 3.5117 & \textbf{4.0149} & 2,550.92 & 2,321.25 & \textbf{2,819.52} \\
& 16 & 64 & 3.5963 & 3.4516 & \textbf{4.0223} & 4,263.54 & 4,292.47 & \textbf{4,792.32} \\
\rowcolor{dmtprow}\cellcolor{white}& 32 & 64 & 3.6094 & 3.4291 & \textbf{4.0556} & 6,259.01 & 6,054.42 & \textbf{7,089.92} \\
\midrule
Agent & 2 & 8 & 4.1367 & 3.9950 & \textbf{4.9991} & 489.04 & 535.73 & \textbf{536.03} \\
& 4 & 8 & 4.2345 & 4.4533 & \textbf{5.2686} & 487.92 & \textbf{567.68} & 564.22 \\
& 8 & 8 & 4.1914 & 4.1486 & \textbf{4.5646} & 688.96 & 562.09 & \textbf{812.00} \\
& 16 & 64 & 4.0647 & 4.4449 & \textbf{5.2731} & 1,104.64 & 1,085.04 & \textbf{1,313.86} \\
\rowcolor{dmtprow}\cellcolor{white}& 32 & 64 & 4.1587 & 4.2493 & \textbf{4.7088} & 1,053.81 & 1,016.94 & \textbf{1,258.44} \\
\bottomrule
\end{tabularx}
\endgroup
\end{table}

Prompt-to-output ratios explain the concurrency trends in Table~\ref{tab:dmtp-batch-sweep}. Code and Math have short prompts and long reasoning outputs, so decoding dominates wall time and DMTP's longer accepted prefixes translate directly into higher throughput at every batch limit ($+11.0\%$ to $+22.7\%$ over DFlash across B2--B32, reaching $+18.77\%$ on Code and $+17.10\%$ on Math at B32). Agent, by contrast, pairs very long prompts with short action responses: at very small batch limits ($B\le 4$, $R=8$), prefill latency, short outputs, and scheduling noise make up a large share of total time and temporarily mask higher acceptance. From B8 onward, KV-heavy decode verification takes a larger share of runtime and DMTP pulls cleanly ahead of both DFlash and Native MTP, delivering $+21.09\%$ (B16) and $+23.75\%$ (B32) higher throughput than DFlash.

In Table~\ref{tab:dmtp-batch-sweep}, B2--B16 run methods on fixed GPU groups with $R=8$ (B2--B8) or $R=64$ (B16), whereas B32 runs DFlash and DMTP back-to-back on the exact same node ($R=64$), making B32 our primary serving reference.

\subsection{First-Candidate Acceptance on Public Short-Context Benchmarks}
\label{app:dmtp-public-first-token}
Complementing the accepted lengths in Table~\ref{tab:dmtp-public}(b), Table~\ref{tab:dmtp-public-t1} reports per-verify first-token ($T_1$) acceptance for DFlash and DMTP(D) on the five public Qwen3.6-27B non-thinking benchmarks (80 conversations per dataset, temperature $1.0$, 1,024-token output cap, target-only verification). Here $T_1$ acceptance is computed at the verification-round level as $1 - H_0 / \sum_{k} H_k$, where $H_0$ counts rounds accepting zero draft tokens (excluding correction or bonus tokens). It measures speculative first-token hit rate per round rather than downstream task accuracy.

\begin{table}[!htbp]
\centering
\caption{\textbf{Per-verify first-token ($T_1$) acceptance (\%) on public Qwen3.6-27B benchmarks (non-thinking, temperature $1.0$).} Rates exclude correction/bonus tokens, and percentage-point gains are computed from unrounded values. Out-of-domain chat benchmarks (Alpaca, MT-Bench) show the largest $T_1$ gains when draft SFT focuses on math and code.}
\label{tab:dmtp-public-t1}
\begingroup
\definecolor{dmtprow}{HTML}{EAF2FC}
\footnotesize
\setlength{\tabcolsep}{8pt}
\renewcommand{\arraystretch}{1.18}
\newcolumntype{R}{>{\raggedleft\arraybackslash}X}
\begin{tabularx}{\linewidth}{@{}l R R R@{}}
\toprule
Dataset & DFlash & DMTP(D) & Gain (pp) \\
\midrule
GSM8K & 79.74 & \textbf{93.83} & +14.08 \\
HumanEval & 90.47 & \textbf{96.26} & +5.79 \\
MATH500 & 81.58 & \textbf{94.80} & +13.21 \\
Alpaca & 66.86 & \textbf{83.45} & +16.59 \\
MT-Bench & 67.27 & \textbf{83.23} & +15.96 \\
\bottomrule
\end{tabularx}
\endgroup
\end{table}

Because OpenPerfectBlend consists mainly of math and code data, standalone DFlash already reaches 90.47\% $T_1$ acceptance on HumanEval, but drops to 66.86\% on Alpaca and 67.27\% on MT-Bench. Reusing frozen native MTP for $T_1$ lifts first-token acceptance by $+5.79$ to $+16.59$ percentage points across all five benchmarks, with the largest jumps on out-of-domain chat tasks (Alpaca and MT-Bench), contributing alongside shifted continuation to DMTP(D)'s accepted-length gains of $+17.17\%$ on MT-Bench and $+21.55\%$ on Alpaca in Table~\ref{tab:dmtp-public}(b).

\subsection{First-Token Survival After Checkpoint Adaptation}
\label{app:dmtp-7k-first-token}
Table~\ref{tab:dmtp-7k-first-token} complements the Qwen3.8-Next accepted lengths in Table~\ref{tab:dmtp-main}(a), comparing open-source DFlash2 against our adapted DMTP checkpoint across six long-context workloads under exact rejection sampling (batch limit 32, temperature $1.0$, 128K input cap, 64K output cap).

\begin{table}[!htbp]
\centering
\caption{\textbf{Per-verify first-token acceptance (\%) after DMTP checkpoint adaptation on Qwen3.8-Next.} Values exclude bonus tokens under exact rejection sampling. Corresponding accepted lengths appear in Table~\ref{tab:dmtp-main}(a).}
\label{tab:dmtp-7k-first-token}
\begingroup
\definecolor{dmtprow}{HTML}{EAF2FC}
\footnotesize
\setlength{\tabcolsep}{8pt}
\renewcommand{\arraystretch}{1.18}
\newcolumntype{R}{>{\raggedleft\arraybackslash}X}
\begin{tabularx}{\linewidth}{@{}l R R R@{}}
\toprule
Workload & DFlash2 & DMTP & Gain (pp) \\
\midrule
General & 61.80 & \textbf{67.40} & +5.60 \\
Reason & 69.46 & \textbf{77.04} & +7.58 \\
Agent & 76.82 & \textbf{82.10} & +5.28 \\
Long-horizon & 74.03 & \textbf{81.43} & +7.40 \\
SWE & 81.96 & \textbf{86.86} & +4.90 \\
Search & 63.23 & \textbf{67.62} & +4.39 \\
\bottomrule
\end{tabularx}
\endgroup
\end{table}

Without retraining DFlash2 from scratch, SFT with dual anchors and a zero-initialized $T_1$ residual raises first-token acceptance by 4.39--7.58 percentage points across all six long-context suites. The overall accepted-length gains in Table~\ref{tab:dmtp-main}(a) combine this stronger MTP first token with adapted prefix-shifted continuation, complementing both the progressive shift ablation in Appendix~\ref{app:dmtp-shift-values} and the matched-$T_1$ CPT study in Appendix~\ref{app:dmtp-cpt}.

\subsection{First-Token Predictors Trained at Scale}
\label{app:dmtp-eagle-first-token}
A natural question is whether DMTP requires a target model pretrained with native MTP, since many open models do not ship with native MTP heads. To test this, we compare native MTP, a single-step EAGLE-style predictor~\citep{li2024eagle,li2025eagle3} trained with cross-entropy, and DFlash trained at matched data scale across six long-context workloads (Table~\ref{tab:dmtp-eagle-first-token}).

\begin{table}[!htbp]
\centering
\caption{\textbf{First-token acceptance (\%) across draft predictors trained at scale.} Per-verify acceptance excluding correction/bonus tokens under matched evaluation settings. EAGLE denotes a single-step predictor trained with cross-entropy.}
\label{tab:dmtp-eagle-first-token}
\begingroup
\footnotesize
\setlength{\tabcolsep}{8pt}
\renewcommand{\arraystretch}{1.18}
\newcolumntype{R}{>{\raggedleft\arraybackslash}X}
\begin{tabularx}{\linewidth}{@{}l R R R@{}}
\toprule
Workload & Native MTP & EAGLE & DFlash \\
\midrule
General & 74.78 & 72.69 & \textbf{75.69} \\
Reason & 75.72 & \textbf{76.31} & 75.91 \\
Agent & 86.40 & 85.06 & \textbf{86.58} \\
Long-horizon & \textbf{83.13} & 80.46 & 83.08 \\
SWE & 86.63 & 87.04 & \textbf{88.08} \\
Search & 80.30 & 80.41 & \textbf{82.83} \\
\bottomrule
\end{tabularx}
\endgroup
\end{table}

Given sufficient training data, both the EAGLE predictor and DFlash reach first-token acceptance comparable to native MTP across all six workloads. This has a practical implication: DMTP's prefix-shifted design is not limited to native-MTP models. A lightweight EAGLE initiator can play the same role by proposing $T_1$ and passing its hidden state into DFlash. Jointly training EAGLE + DFlash during continued pretraining (CPT) offers a practical way to bring cooperative block drafting to any target LLM without pretraining-scale cost. Evaluating the latency, OOD generalization, and training cost of an EAGLE + DFlash system is an exciting avenue for future work.
\FloatBarrier

\section{Numerical Data Underlying the Figures}
\label{app:dmtp-figure-data}
For reproducibility and exact comparison, this section lists all numerical values plotted in the main paper's figures. Position-wise acceptance rates are per-verify percentages excluding correction or bonus tokens, whereas accepted lengths include the target model's correction or bonus token. Exact values for the progressive shift ablation (Figure~\ref{fig:dmtp-shift-motivation}), latency breakdown (Figure~\ref{fig:dmtp-latency}), and serving batch sweep (Figure~\ref{fig:dmtp-serving}) appear in Tables~\ref{tab:dmtp-shift-values}, \ref{tab:dmtp-latency-breakdown}, and~\ref{tab:dmtp-batch-sweep}, respectively.

\subsection{First-Token and Cumulative Acceptance}
\label{app:dmtp-figure-first}
Table~\ref{tab:dmtp-first-source} lists the standalone first-token acceptance rates plotted in Figure~\ref{fig:dmtp-first-token} on Qwen3.6-35B-A3B across six workloads at temperature $1.0$, showing native MTP's consistent 10.50--14.43 percentage-point lead over DFlash. Because standalone native MTP and closed-loop DMTP rollouts visit slightly different generated prefixes, native MTP's standalone rates in Table~\ref{tab:dmtp-first-source} differ slightly from DMTP's $T_1$ rates in Table~\ref{tab:dmtp-cumulative-source}.

\begin{table}[htbp]
\centering
\caption{\textbf{Numerical data for Figure~\ref{fig:dmtp-first-token} (standalone first-token acceptance, \%).} Evaluated on identical Qwen3.6-35B-A3B prompt sets at temperature $1.0$, excluding correction/bonus tokens.}
\label{tab:dmtp-first-source}
\begingroup
\definecolor{dmtprow}{HTML}{EAF2FC}
\footnotesize
\setlength{\tabcolsep}{6pt}
\renewcommand{\arraystretch}{1.18}
\newcolumntype{R}{>{\raggedleft\arraybackslash}X}
\begin{tabularx}{\linewidth}{@{}l R R R R R R@{}}
\toprule
Method & Code & Math & MT-Bench & Agent & SWE & Long-horizon \\
\midrule
DFlash & 71.00 & 73.73 & 61.36 & 79.53 & 59.33 & 75.13 \\
\rowcolor{dmtprow}Native MTP & \textbf{82.32} & \textbf{84.23} & \textbf{74.78} & \textbf{90.17} & \textbf{73.76} & \textbf{86.65} \\
\bottomrule
\end{tabularx}
\endgroup
\end{table}

Table~\ref{tab:dmtp-cumulative-source} lists the cumulative acceptance curves $s_i = \Pr(N\geq i)$ plotted in Figure~\ref{fig:dmtp-main-position}, computed from pooled verification histograms over the seven draft positions ($T_1$--$T_7$). Unlike the conditional rates $c_i$ in Table~\ref{tab:dmtp-conditional}, each value reflects survival through all earlier positions $1,\ldots,i-1$. Across all six workloads and all seven positions (42 of 42 entries), DMTP maintains a clear lead over standalone DFlash, showing how a stronger $T_1$ and $T_1$-conditioned continuation reinforce each other across the block.

\begin{table}[htbp]
\centering
\caption{\textbf{Numerical data for Figure~\ref{fig:dmtp-main-position} (cumulative prefix acceptance, \%).} Per-verify fraction of rounds accepting all candidates through position $T_i$ ($K=7$), excluding correction/bonus tokens.}
\label{tab:dmtp-cumulative-source}
\begingroup
\definecolor{dmtprow}{HTML}{EAF2FC}
\footnotesize
\setlength{\tabcolsep}{5pt}
\renewcommand{\arraystretch}{1.15}
\newcolumntype{R}{>{\raggedleft\arraybackslash}X}
\begin{tabularx}{\linewidth}{@{}ll R R R R R R R@{}}
\toprule
Workload & Method & $T_1$ & $T_2$ & $T_3$ & $T_4$ & $T_5$ & $T_6$ & $T_7$ \\
\midrule
Code & DFlash & 71.00 & 49.97 & 35.79 & 25.34 & 17.85 & 12.66 & 8.98 \\
\rowcolor{dmtprow}\cellcolor{white}& DMTP & \textbf{81.81} & \textbf{62.85} & \textbf{46.73} & \textbf{34.31} & \textbf{25.05} & \textbf{18.29} & \textbf{13.13} \\
\midrule
Math & DFlash & 73.73 & 53.77 & 39.69 & 29.32 & 21.51 & 15.46 & 10.92 \\
\rowcolor{dmtprow}\cellcolor{white}& DMTP & \textbf{83.72} & \textbf{66.01} & \textbf{50.34} & \textbf{38.44} & \textbf{29.08} & \textbf{21.67} & \textbf{15.74} \\
\midrule
MT-Bench & DFlash & 61.36 & 38.07 & 26.02 & 19.09 & 14.58 & 11.10 & 8.31 \\
\rowcolor{dmtprow}\cellcolor{white}& DMTP & \textbf{73.72} & \textbf{49.32} & \textbf{33.43} & \textbf{24.41} & \textbf{18.61} & \textbf{14.29} & \textbf{10.68} \\
\midrule
Agent & DFlash & 79.53 & 65.09 & 54.68 & 46.95 & 40.87 & 35.48 & 30.06 \\
\rowcolor{dmtprow}\cellcolor{white}& DMTP & \textbf{86.98} & \textbf{74.91} & \textbf{64.06} & \textbf{55.67} & \textbf{48.83} & \textbf{42.65} & \textbf{36.76} \\
\midrule
SWE & DFlash & 59.33 & 38.12 & 26.07 & 19.27 & 15.35 & 12.43 & 10.55 \\
\rowcolor{dmtprow}\cellcolor{white}& DMTP & \textbf{71.79} & \textbf{52.11} & \textbf{38.89} & \textbf{31.00} & \textbf{26.17} & \textbf{22.26} & \textbf{18.30} \\
\midrule
Long-horizon & DFlash & 75.13 & 57.73 & 46.28 & 37.03 & 30.70 & 25.69 & 21.69 \\
\rowcolor{dmtprow}\cellcolor{white}& DMTP & \textbf{80.56} & \textbf{64.97} & \textbf{51.97} & \textbf{43.25} & \textbf{36.78} & \textbf{31.67} & \textbf{27.16} \\
\bottomrule
\end{tabularx}
\endgroup
\end{table}

\FloatBarrier
\subsection{Hidden-State Interfaces and Anchor Layouts}
\label{app:dmtp-figure-interfaces}
Tables~\ref{tab:dmtp-hidden-source} and~\ref{tab:dmtp-anchor-source} report the exact numbers behind the architectural ablations in Figure~\ref{fig:dmtp-ablation-design}(a--c). To match the main paper's protocol, every entry reports pooled per-verify accepted length (including correction/bonus tokens) recomputed from the eight-request verification histograms ($K=7$).

In Table~\ref{tab:dmtp-hidden-source}, all variants except original DFlash use single-anchor $T_1$ conditioning so that only the hidden-state interface changes. Specifically, \emph{candidate conditioning} feeds $[T_1,\mathrm{MASK}^6]$ without MTP hidden states, while \emph{token-hidden fusion at $T_1$} normalizes, concatenates, and projects $E(T_1)$ and $h_{\mathrm{mtp}}$ to replace the $T_1$ embedding. \emph{MTP hidden residual at $T_1$} adds a projected residual only to $E(T_1)$ while leaving mask embeddings untouched, whereas \emph{MTP residual at six MASK slots} broadcasts the residual across all six mask positions (evaluated on identical prompts via an auxiliary path). Finally, \emph{target-MTP context fusion} concatenates five target feature layers with MTP features through a $6H \to H$ projection, and \emph{MTP-only context} replaces target KV context entirely with MTP features. Dashes denote configurations without a full eight-request run.

\begin{table}[htbp]
\centering
\caption{\textbf{Cross-domain hidden-state interface ablation (numerical data for Figure~\ref{fig:dmtp-ablation-design}a).} Per-verify accepted length across eight requests ($K=7$), including correction/bonus tokens. ``--'' indicates no complete eight-request run on that subset.}
\label{tab:dmtp-hidden-source}
\begingroup
\definecolor{dmtprow}{HTML}{EAF2FC}
\footnotesize
\setlength{\tabcolsep}{6pt}
\renewcommand{\arraystretch}{1.18}
\newcolumntype{R}{>{\raggedleft\arraybackslash}X}
\begin{tabularx}{\linewidth}{@{}l R R R@{}}
\toprule
Representation interface & Code & Math & MT-Bench \\
\midrule
Original DFlash & 3.3348 & 3.5858 & 2.8315 \\
Candidate conditioning, no MTP hidden (ref) & 3.7574 & 3.8652 & \textbf{3.3494} \\
Token-hidden fusion at $T_1$ & \textbf{3.9480} & \textbf{3.9055} & 3.2105 \\
\rowcolor{dmtprow}MTP hidden residual at $T_1$ (DMTP) & 3.8492 & 3.8593 & 3.2782 \\
MTP residual at six MASK slots (auxiliary run) & 3.8318 & -- & -- \\
Target-MTP context fusion ($6H\to H$) & 3.6954 & -- & -- \\
MTP-only context & 3.5469 & 3.7272 & -- \\
\bottomrule
\end{tabularx}
\endgroup
\end{table}

Table~\ref{tab:dmtp-anchor-source} compares single-anchor $(T_1,\mathrm{MASK}^6)$ and dual-anchor $(A,T_1,\mathrm{MASK}^6)$ inputs under both $T_1$ residual injection and full token-hidden fusion. While both layouts reach similar accuracy when trained from scratch, keeping both $A$ and $T_1$ with a lightweight $T_1$ residual consistently improves acceptance across all three domains ($+3.41\%$ on Code, $+4.31\%$ on Math, and $+2.10\%$ on MT-Bench), and enriches $E(T_1)$ with extra MTP hidden cues without overwriting the token embedding. We therefore adopt dual anchors with a zero-initialized $T_1$ residual as DMTP's default design.

\begin{table}[htbp]
\centering
\caption{\textbf{Single-anchor versus dual-anchor layouts (numerical data for Figure~\ref{fig:dmtp-ablation-design}b,c).} Per-verify accepted length ($K=7$) comparing single-anchor $(T_1,\mathrm{MASK}^6)$ and dual-anchor $(A,T_1,\mathrm{MASK}^6)$ inputs under $T_1$ residual injection or full token-hidden fusion.}
\label{tab:dmtp-anchor-source}
\begingroup
\footnotesize
\setlength{\tabcolsep}{6pt}
\renewcommand{\arraystretch}{1.18}
\newcolumntype{R}{>{\raggedleft\arraybackslash}X}
\begin{tabularx}{\linewidth}{@{}l R R R R@{}}
\toprule
& \multicolumn{2}{c}{$T_1$ Residual} & \multicolumn{2}{c}{Token-Hidden Fusion} \\
\cmidrule(lr){2-3}\cmidrule(l){4-5}
Workload & Single ($T_1$) & Dual ($A,T_1$) & Single ($T_1$) & Dual ($A,T_1$) \\
\midrule
Code & 3.4072 & \textbf{3.5234} & 3.5051 & \textbf{3.5082} \\
Math & 3.6522 & \textbf{3.8095} & \textbf{3.8126} & 3.6301 \\
MT-Bench & 2.7921 & \textbf{2.8507} & 2.8345 & \textbf{3.0591} \\
\bottomrule
\end{tabularx}
\endgroup
\end{table}

\FloatBarrier
\subsection{Terminal Bench Joint-Training Curves}
\label{app:dmtp-figure-joint}
Table~\ref{tab:dmtp-joint-source} provides the exact position-wise cumulative acceptance values plotted in Figures~\ref{fig:dmtp-joint-summary} and~\ref{fig:dmtp-joint} on the 255-request Terminal Bench evaluation set ($K=8$, exact rejection sampling, concurrency 32). The percentage-point gain curves subtract the frozen-MTP baseline row from each joint-training row, showing consistent improvements across all positions $T_1$--$T_8$ at the matched training budget and after extended joint training.

\begin{table}[htbp]
\centering
\caption{\textbf{Numerical data for the Terminal Bench joint-training curves (Figures~\ref{fig:dmtp-joint-summary} and~\ref{fig:dmtp-joint}).} Entries report per-verify cumulative acceptance (\%, excluding correction/bonus tokens) at $K=8$, along with per-verify accepted length $\tau$ (including bonus).}
\label{tab:dmtp-joint-source}
\begingroup
\definecolor{dmtprow}{HTML}{EAF2FC}
\footnotesize
\setlength{\tabcolsep}{4pt}
\renewcommand{\arraystretch}{1.18}
\newcolumntype{R}{>{\raggedleft\arraybackslash}X}
\begin{tabularx}{\linewidth}{@{}l R R R R R R R R R@{}}
\toprule
Setting & $\tau$ & $T_1$ & $T_2$ & $T_3$ & $T_4$ & $T_5$ & $T_6$ & $T_7$ & $T_8$ \\
\midrule
Frozen MTP & 4.3178 & 79.80 & 63.99 & 50.43 & 39.68 & 31.77 & 26.01 & 21.73 & 18.37 \\
Joint (matched budget) & 4.4239 & 81.18 & 65.62 & 51.65 & 41.13 & 33.19 & 27.32 & 22.93 & 19.37 \\
\rowcolor{dmtprow}Joint (extended training) & \textbf{4.4745} & \textbf{81.45} & \textbf{66.11} & \textbf{52.49} & \textbf{41.96} & \textbf{33.94} & \textbf{28.07} & \textbf{23.53} & \textbf{19.89} \\
\bottomrule
\end{tabularx}
\endgroup
\end{table}
\FloatBarrier

\section{Position-Wise Survival Under Matched First-Token Accuracy}
\label{app:dmtp-cpt}
To test whether prefix-shifted parallel continuation still helps when a parallel drafter already matches or exceeds native MTP on the first token, we initialize DMTP from a continued-pretrained (CPT) DFlash2 checkpoint and fine-tune only the draft layers, selector, and zero-initialized $T_1$ residual using SFT ($K=8$, target and MTP frozen, Section~\ref{sec:dmtp-ckpt}). Table~\ref{tab:dmtp-cpt-full} reports per-verify cumulative acceptance $\Pr(N\geq i)$ across all eight draft positions $T_1$--$T_8$ on six long-context workloads.

Even though DFlash2 CPT matches or beats native MTP at $T_1$ on five of the six workloads (e.g., 88.08\% vs.\ 86.63\% on SWE and 82.83\% vs.\ 80.30\% on Search), conditioning parallel continuation on MTP's sampled $T_1$ lets DMTP overtake DFlash2 CPT starting at $T_2$ and lead at every continuation position $T_2$--$T_8$ across all six workloads (42 of 42 entries). Notably, DMTP also surpasses multi-step sequential MTP across $T_2$--$T_8$ on General, Agent, Long-horizon, and SWE, and at $T_2$--$T_3$ on Reason and Search, further confirming the structural benefit of prefix shifting shown in Section~\ref{sec:dmtp-ablation}.

\begin{table}[htbp]
\centering
\caption{\textbf{Position-wise cumulative acceptance (\%) under matched first-token accuracy ($K=8$, adaptation from DFlash2 CPT).} Values exclude bonus tokens. Even when DFlash2 CPT matches or exceeds native MTP at $T_1$, shifted continuation in DMTP beats DFlash2 CPT at every continuation position ($T_2$--$T_8$) across all six workloads and surpasses sequential MTP on most positions.}
\label{tab:dmtp-cpt-full}
\begingroup
\definecolor{dmtprow}{HTML}{EAF2FC}
\footnotesize
\setlength{\tabcolsep}{4pt}
\renewcommand{\arraystretch}{1.16}
\newcolumntype{R}{>{\raggedleft\arraybackslash}X}
\begin{tabularx}{\linewidth}{@{}ll R R R R R R R R@{}}
\toprule
Dataset & Method & $T_1$ & $T_2$ & $T_3$ & $T_4$ & $T_5$ & $T_6$ & $T_7$ & $T_8$ \\
\midrule
General & MTP & 74.78 & 55.27 & 41.38 & 31.72 & 24.86 & 19.61 & 15.53 & 12.22 \\
& DFlash2 CPT & \textbf{75.69} & 54.31 & 39.28 & 29.14 & 22.36 & 17.65 & 14.21 & 11.68 \\
\rowcolor{dmtprow}\cellcolor{white}& DMTP & 75.15 & \textbf{57.90} & \textbf{43.99} & \textbf{34.07} & \textbf{27.16} & \textbf{21.99} & \textbf{18.05} & \textbf{14.88} \\
\midrule
Reason & MTP & 75.72 & 56.45 & 42.00 & \textbf{31.50} & \textbf{23.75} & \textbf{18.00} & \textbf{13.74} & \textbf{10.53} \\
& DFlash2 CPT & 75.91 & 52.58 & 35.74 & 24.60 & 17.39 & 12.65 & 9.43 & 7.21 \\
\rowcolor{dmtprow}\cellcolor{white}& DMTP & \textbf{76.04} & \textbf{58.73} & \textbf{43.00} & 30.93 & 22.35 & 16.36 & 12.20 & 9.26 \\
\midrule
Agent & MTP & 86.40 & 74.73 & 65.12 & 57.18 & 50.52 & 44.53 & 39.39 & 34.58 \\
& DFlash2 CPT & \textbf{86.58} & 73.56 & 63.10 & 55.19 & 48.88 & 43.91 & 39.73 & 36.21 \\
\rowcolor{dmtprow}\cellcolor{white}& DMTP & 85.87 & \textbf{75.17} & \textbf{65.61} & \textbf{57.75} & \textbf{51.42} & \textbf{46.21} & \textbf{41.97} & \textbf{38.31} \\
\midrule
Long-horizon & MTP & \textbf{83.13} & 70.00 & 59.28 & 50.99 & 44.08 & 38.03 & 33.00 & 28.76 \\
& DFlash2 CPT & 83.08 & 66.99 & 54.66 & 45.35 & 38.24 & 32.81 & 28.70 & 25.29 \\
\rowcolor{dmtprow}\cellcolor{white}& DMTP & 82.87 & \textbf{71.35} & \textbf{60.92} & \textbf{51.96} & \textbf{44.69} & \textbf{39.03} & \textbf{34.55} & \textbf{30.72} \\
\midrule
SWE & MTP & 86.63 & 74.61 & 64.64 & 56.03 & 49.13 & 43.12 & 37.71 & 32.85 \\
& DFlash2 CPT & \textbf{88.08} & 74.54 & 62.72 & 53.24 & 45.93 & 40.39 & 35.98 & 32.56 \\
\rowcolor{dmtprow}\cellcolor{white}& DMTP & 86.51 & \textbf{76.54} & \textbf{66.35} & \textbf{57.43} & \textbf{49.85} & \textbf{44.03} & \textbf{39.37} & \textbf{35.53} \\
\midrule
Search & MTP & 80.30 & 62.97 & 49.32 & \textbf{38.90} & \textbf{30.69} & \textbf{24.50} & \textbf{19.57} & \textbf{15.57} \\
& DFlash2 CPT & \textbf{82.83} & 64.08 & 48.40 & 36.71 & 28.27 & 22.17 & 17.76 & 14.45 \\
\rowcolor{dmtprow}\cellcolor{white}& DMTP & 79.81 & \textbf{64.93} & \textbf{50.41} & 38.55 & 29.67 & 23.36 & 18.80 & 15.25 \\
\bottomrule
\end{tabularx}
\endgroup
\end{table}

\section{Open-Source Artifacts}
\label{app:dmtp-resources}
To support reproduction, benchmarking, and production deployment, we release our trained DMTP checkpoints, serving code, training recipes, and companion technical notes:
\begin{itemize}[leftmargin=5mm,itemsep=4pt,topsep=3pt]
    \item \textbf{Qwen3.6-27B DMTP checkpoint:} \href{https://huggingface.co/Piggy12/Qwen3.6-27B-DMTP}{\texttt{Piggy12/Qwen3.6-27B-DMTP}} (dual-anchor DMTP trained from scratch on OpenPerfectBlend with frozen target and native MTP).
    \item \textbf{Qwen3.8-Next-120B DMTP checkpoint:} \href{https://huggingface.co/Piggy12/Qwen3.8-next-120b-dmtp}{\texttt{Piggy12/Qwen3.8-next-120b-dmtp}} (adapted from open-source DFlash2 via dual-anchor zero-initialized $T_1$ residual fine-tuning).
    \item \textbf{SGLang serving engine:} \href{https://github.com/Sylvan820/sglang-dmtp}{\texttt{Sylvan820/sglang-dmtp}} and upstream \href{https://github.com/sgl-project/sglang/pull/50270}{SGLang PR \#50270}, implementing temporary FlashAttention-3 page tables, fixed eight-slot MTP KV state management, fused $T_1$ CUDA Graphs, and sparse rejection sampling.
    \item \textbf{SpecForge training framework:} \href{https://github.com/sgl-project/SpecForge/pull/1067}{SpecForge PR \#1067}, supporting frozen-MTP from-scratch training, rapid checkpoint adaptation, and joint LogTV MTP-draft fine-tuning.
    \item \textbf{Companion technical blog:} \href{https://sylvan820.github.io/D-MTP/}{\texttt{sylvan820.github.io/D-MTP/}}, providing development logs, architectural notes, and extra rollout visualizations.
\end{itemize}

\FloatBarrier
\section{Extended Joint MTP-Draft Training Analysis}
\label{app:dmtp-joint-figure}
When a domain-adapted parallel drafter matches or beats frozen native MTP on the first token, keeping MTP frozen limits DMTP's upside. Section~\ref{sec:dmtp-joint} resolves this via joint MTP-draft SFT with the decoupled LogTV and selector-KL loss (Equations~\ref{eq:dmtp-logtv}--\ref{eq:dmtp-joint}). In our implementation, per-block losses are averaged over active token masks alongside any native MoE auxiliary losses. No selector or cross-entropy loss is placed on $T_1$, and $h_{\mathrm{mtp}}$ stays detached ($\operatorname{sg}(h_{\mathrm{mtp}})$) so that continuation and selector gradients do not flow into MTP. The $T_1$ LogTV term supervises MTP, with applicable native MoE auxiliary losses retained.

\begin{figure}[!htbp]
\centering
\includegraphics[width=\linewidth]{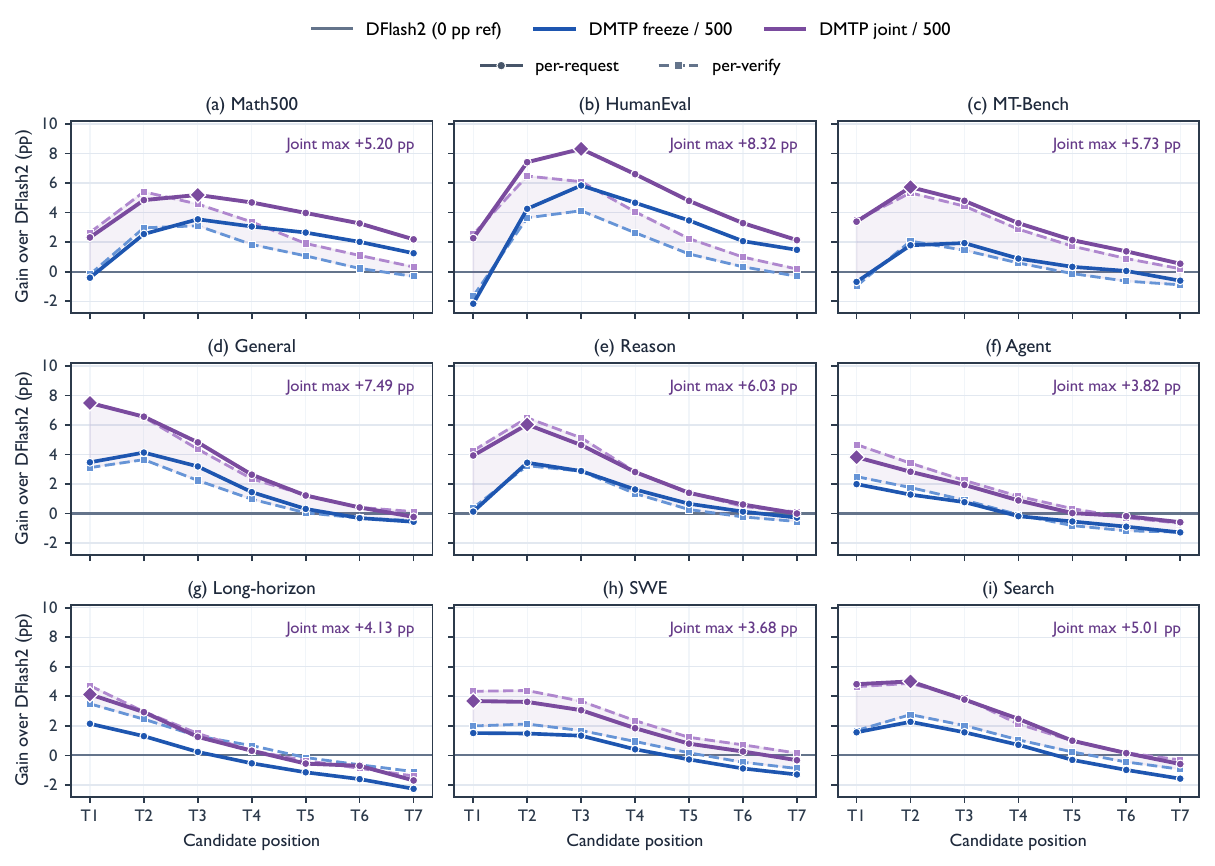}
\caption{\textbf{Cross-benchmark position-wise cumulative acceptance gain over DFlash2 (pp) after SFT: frozen MTP versus joint training.} Blue curves show frozen-MTP DMTP, and purple curves show joint MTP-draft fine-tuning (dashed lines show per-verify results, and solid lines show per-request results). Joint training turns frozen-MTP's slight $T_1$ deficits on Math500, HumanEval, and MT-Bench into positive gains and boosts acceptance across later positions on all nine benchmarks.}
\label{fig:dmtp-joint-cross}
\end{figure}

Figure~\ref{fig:dmtp-joint-cross} evaluates this recipe across nine benchmarks (Math500, HumanEval, MT-Bench, General, Reason, Agent, Long-horizon, SWE, and Search) after adaptation, plotting position-wise cumulative acceptance gains (in percentage points) over standalone DFlash2. Looking at the per-verify dashed curves, frozen-MTP DMTP starts slightly behind DFlash2 at $T_1$ on Math500, HumanEval, and MT-Bench before pulling ahead at $T_2$--$T_4$ through prefix shifting. Unfreezing MTP under joint LogTV training turns the per-verify $T_1$ gain positive on all three benchmarks and lifts acceptance across nearly every downstream position on all nine tasks (peaking at $+3.68$ to $+8.32$ percentage points).

\begin{figure}[!htbp]
\centering
\includegraphics[width=\linewidth]{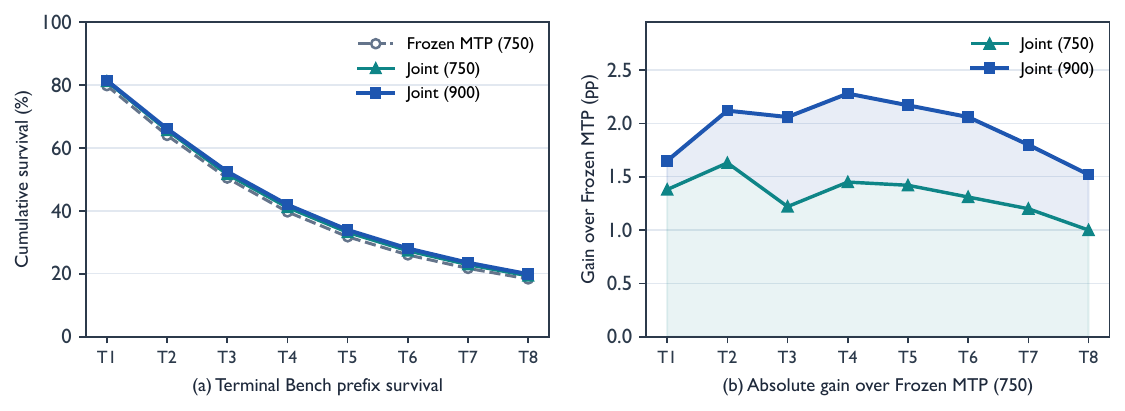}
\caption{\textbf{Position-wise prefix acceptance on Terminal Bench under frozen-MTP versus joint SFT ($K=8$, rejection sampling).} Left: cumulative acceptance across $T_1$--$T_8$ (excluding bonus tokens). Right: percentage-point gain over frozen-MTP adaptation. Joint fine-tuning improves acceptance across every position at a matched training budget, with further gains from extended joint training.}
\label{fig:dmtp-joint}
\end{figure}

Figure~\ref{fig:dmtp-joint} shows the full position-wise curves ($T_1$--$T_8$) for the Terminal Bench offline SFT study in Figure~\ref{fig:dmtp-joint-summary}. Starting from the same CPT checkpoint with the target model frozen, both runs train offline on collected Terminal Bench traces (rather than online RL) and are evaluated on 255 requests under exact rejection sampling ($K=8$, concurrency 32). At a matched training budget, joint training raises cumulative acceptance at every position $T_1$--$T_8$ by $+1.00$ to $+1.63$ percentage points (lifting $T_1$ from 79.80\% to 81.18\% and accepted length by $+2.46\%$). Extending joint training widens the gain to $+1.52$ to $+2.28$ percentage points across all eight positions ($+3.63\%$ accepted length, with exact values in Table~\ref{tab:dmtp-joint-source}). Together, Figures~\ref{fig:dmtp-joint-cross} and~\ref{fig:dmtp-joint} show that prefix shifting and joint first-token adaptation work hand in hand: prefix shifting improves parallel continuation, while joint SFT restores a clear first-token lead on specialized domains.

\FloatBarrier
\section{Detailed Related Work}
\label{app:dmtp-related-detailed}
This section expands Section~\ref{sec:dmtp-related} across eight topics: exact speculative sampling, parallel block drafters, intra-block causality, native MTP and feature-level drafting, acceptance-aligned losses, serving-aware verification, RL rollout acceleration, and multimodal speculative decoding. Reviewing each line of work clarifies which bottlenecks DMTP addresses directly and how it complements orthogonal techniques.

\subsection{Exact Speculative Sampling and Prefix Survival}
\label{app:dmtp-related-exact}
\citet{leviathan2023fast} and \citet{chen2023accelerating} introduced exact speculative decoding by pairing fast multi-token drafting with single-pass target verification and rejection resampling, guaranteeing exact target-distribution recovery for any draft model. While output quality is lossless regardless of draft accuracy, verification stops at the first rejected token, making early-position accuracy the main driver of speedup. Tree-based methods such as SpecInfer~\citep{specinfer} and EAGLE-2~\citep{li2024eagle2} widen candidate coverage through static or confidence-guided branching. DMTP instead strengthens single-chain acceptance directly: native MTP supplies a stronger first token, and parallel continuation conditions on that chosen token. Because DMTP changes only how the draft block is proposed, it keeps target verification untouched and works under both target-only and exact rejection sampling.

\subsection{Parallel and Block-Diffusion Drafters}
\label{app:dmtp-related-parallel}
To avoid the $O(K)$ sequential latency of autoregressive drafters, parallel and block-diffusion drafters predict an entire block in one forward pass. PARD~\citep{pard} adapts small autoregressive drafters into parallel mask predictors, while P-EAGLE~\citep{peagle} turns feature-level autoregression into parallel multi-token prediction. DFlash~\citep{chen2026dflash} conditions a lightweight bidirectional block drafter on multi-layer target KV features, DFlare~\citep{dflare} scales parallel drafting via layer-wise feature fusion, and DFlash2~\citep{dflash2} improves later positions with block-local convolutions, a path selector, and adaptive continuation losses. DMTP is orthogonal to and composable with existing MTP and parallel-drafting optimizations: instead of replacing these parallel drafters or their training recipes, it runs single-step native MTP ahead of them, shifts their first unknown position from $T_1$ to $T_2$, and passes $h_{\mathrm{mtp}}$ via a zero-initialized residual, plugging cleanly into both DFlash and DFlash2.

\subsection{Intra-Block Causality and Semi-Autoregressive Drafting}
\label{app:dmtp-related-causal}
Standard parallel drafters predict all $K$ positions at once without intra-block token dependencies, leaving uncertainty about earlier tokens and causing accuracy to decay across the block. Domino~\citep{domino} adds a causal refinement head under a two-stage curriculum, while DSpark~\citep{cheng2026dspark} uses a Markov or RNN head with confidence-based verification scheduling. Like DFlash2's block-local improvements, these methods primarily strengthen dependencies within the draft. DMTP instead focuses on first-token accuracy and OOD generalization by reusing native MTP, then shifts DFlash's prediction window to start at $T_2$. Every mask position can thus attend to the sampled $T_1$ and its hidden state. Our ablation (Section~\ref{sec:dmtp-ablation}) separates first-token replacement, hidden-state injection, and prefix shifting, showing substantial shift gains even at similar first-token accuracy. This \emph{pre-backbone} design complements block-internal refinement. Compatibility is demonstrated with DFlash and DFlash2, while combinations with Domino or DSpark remain future work.

\subsection{Native MTP and Feature-Conditioned Drafting}
\label{app:dmtp-related-mtp}
Native Multi-Token Prediction~\citep{gloeckle2024bmtp} trains auxiliary future-token heads alongside the target trunk during pretraining, as adopted in DeepSeek-V3~\citep{deepseekv3} and Qwen3.5~\citep{qwen35}. Pretraining at full corpus scale gives native MTP broad domain coverage and strong first-token accuracy without a separate full draft model. Post-hoc feature-level drafters also use target representations: Medusa~\citep{medusa} uses independent decoding heads, EAGLE~\citep{li2024eagle} autoregresses over target features with token embeddings, EAGLE-3~\citep{li2025eagle3} adds multi-layer feature fusion and multi-step training, and KVShot~\citep{liu2026kvshot} reuses target KV caches to reduce feature drift. Native MTP's strength comes with a limitation: multi-step rollout requires sequential execution, increasing latency with block length. DMTP uses MTP only to obtain $T_1$ and $h_{\mathrm{mtp}}$, then delegates continuation to DFlash. Its target KV path stays unchanged, while $h_{\mathrm{mtp}}$ enters through a residual on $E(T_1)$. This combines MTP's pretrained first-token capabilities with DFlash's parallel efficiency, with prefix shifting providing gains even when first-token accuracies are close.

\subsection{Draft Distillation and Acceptance-Aligned Objectives}
\label{app:dmtp-related-training}
Beyond model architecture, draft alignment benefits substantially from better training data and acceptance-aware losses. DistillSpec~\citep{distillspec} shows that logit distillation outperforms hard-label cross-entropy, while Generalized Knowledge Distillation~\citep{gkd}, Draft-OPD~\citep{draftopd}, and EAGLE-3~\citep{li2025eagle3} train on draft-generated prefixes to reduce exposure bias during closed-loop decoding. To remove the sequential rollout bottleneck of conventional on-policy pipelines for parallel drafters, SpecOPD~\citep{specopd} parallelizes draft sampling and target supervision across branch-parallel proposal trees rooted at reference anchors, complementing general LLM training-efficiency techniques that reduce adapter memory footprints~\citep{loram} and gradient communication overhead~\citep{edgc}. Because verification stops at the first rejection, recent losses also replace uniform cross-entropy with position-decayed weights (D-PACE~\citep{dpace}), expected decoding rounds~\citep{edr}, or direct distribution-overlap objectives such as LK Losses~\citep{lklosses} and end-to-end TV acceptance surrogates~\citep{bebop,specopd}. Our joint SFT recipe builds directly on this overlap perspective, minimizing position-decayed LogTV across $T_1$--$T_8$ alongside top-16 selector KL while detaching $h_{\mathrm{mtp}}$ to separate first-token and continuation gradients. Extending our teacher-forced frozen-MTP recipe to on-policy MTP-sampled $T_1$ inputs is a natural next step.

\subsection{Verification Scheduling and Serving Efficiency}
\label{app:dmtp-related-serving}
Turning higher accepted lengths into real serving speedups depends on batch size, KV memory traffic, and verification scheduling. Dynamic methods such as EAGLE-2~\citep{li2024eagle2}, D-cut~\citep{liu2026dcut}, and DSpark~\citep{cheng2026dspark} prune draft trees or adjust verification lengths based on confidence and live server load. Our batch-limit sweep in Appendix~\ref{app:dmtp-batch-sweep} shows why system co-design matters: on long-prompt Agent tasks at low concurrency ($B\le 4$), prefill and scheduling overhead can hide longer accepted prefixes, whereas from $B=8$ onward ($B\in\{8, 16, 32\}$), our fused CUDA-Graph runtime turns longer accepted prefixes into 17.10\%--23.75\% end-to-end throughput gains in SGLang~\citep{sglang}. Pairing DMTP with load-aware dynamic block sizing could improve serving efficiency further. At the model-serving layer, complementary system techniques address weight memory and sharing, such as hierarchical multi-tenant knowledge sharing in HMI~\citep{hmi} and on-the-fly MoE expert loading on memory-constrained GPUs in FloE~\citep{floe}, which can be composed with DMTP's serving runtime.

\subsection{RL Rollouts and Joint Domain Adaptation}
\label{app:dmtp-related-rl}
Speculative decoding is increasingly used to speed up long-horizon reinforcement learning (RL) rollouts, where generation dominates training time. SPEC-RL~\citep{liu2025specrl}, SRT~\citep{chang2026srt}, and RL-speculative engines~\citep{iso2026rlspec} accelerate rollouts via trajectory reuse and tree drafting, while BeBop~\citep{bebop} studies MTP drift under RL and Aurora~\citep{wang2026aurora} updates drafters online from live traces as the target policy evolves. Our Terminal Bench study (Section~\ref{sec:dmtp-joint-results} and Appendix~\ref{app:dmtp-joint-figure}) focuses on an offline step toward this setting: jointly fine-tuning MTP and DFlash2 on collected RL traces with the target model frozen. Showing that joint adaptation lifts acceptance across all eight positions on specialized agentic traces provides a clean foundation for future online RL loops.

\subsection{Multimodal Speculative Decoding and Cache Compression}
\label{app:dmtp-related-multimodal}
Extending speculative decoding to vision-language and multimodal models introduces distinct visual encoding, prefilling, and long-context cache bottlenecks~\citep{vlmefficiency}. SpecVLM~\citep{ji2025specvlm} prunes redundant video tokens on the draft side while retaining full-target verification to preserve lossless outputs, whereas LVSpec~\citep{ji2026lvspec} explores visual-semantic guidance for loosely speculative decoding when exact token-level equivalence can be relaxed. Complementary acceleration and compression methods reduce multimodal conditioning, KV-cache, and iterative decoding overhead via skill-driven token budget allocation in OmniDelta~\citep{huang2026omnidelta}, head-specific hybrid KV compression in HybridKV~\citep{hybridkv}, hybrid KV compression for autoregressive video diffusion in Forcing-KV~\citep{ji2026forcingkv}, and adaptive action reuse to reduce denoising steps in vision-language-action models~\citep{vlarolling}. Extending DMTP's cooperative MTP initiation and shifted parallel continuation to multimodal targets, and combining it with draft-side visual pruning or KV compression while preserving exact target verification, is a promising avenue for future work.

\end{document}